\documentclass[a4paper,fleqn]{cas-dc}

\usepackage[numbers]{natbib}

\usepackage{amsmath,amssymb,amsfonts}
\usepackage{algorithmic}
\usepackage{algorithm}
\usepackage{array}
\usepackage{textcomp}
\usepackage{stfloats}
\usepackage{verbatim}
\usepackage{graphicx}
\usepackage{booktabs}
\input{lc_commands.tex}
\usepackage{multirow}
\usepackage{pifont}
\newcommand{\mone}{PMMM}
\newcommand{\model}{PCMM}
\newcommand{\meta}{C2C}
\usepackage{makecell}
\newtheorem{definition}{\textbf{Definition}}[section]
\usepackage[caption=false,font=normalsize,labelfont=sf,textfont=sf]{subfig}

\usepackage{hyperref}
\hypersetup{hidelinks,
	colorlinks=true,
	allcolors=black,
	pdfstartview=Fit,
	breaklinks=true}

 \usepackage{url}

\begin{document}
\let\WriteBookmarks\relax
\def\floatpagepagefraction{1}
\def\textpagefraction{.001}

\shorttitle{Meta-multigraph search}

\shortauthors{Li et~al.}

\title [mode = title]{Meta-multigraph Search: Rethinking Meta-structure on Heterogeneous Information Networks}

\author[1,2]{Chao Li}
\cormark[1]
\ead{D201880880@hust.edu.cn}
\credit{Conceptualization of this study, Methodology, Software, Writing, and revision}

\affiliation[1]{
    organization={School of Computer Science and Technology, Huazhong University of Science and Technology},
    city={Wuhan},
    postcode={430074}, 
    country={China}}

\author[1,2]{Hao Xu}
\ead{M202173589@hust.edu.cn}
\credit{Software, Writing, and revision}
\author[1,2]{Kun He}
[orcid=0000-0001-7627-4604]
\ead{brooklet60@hust.edu.cn}
\credit{Conceptualization of this study,
Methodology, Supervision, Writing, and revision}

\cormark[2]


\affiliation[2]{
    organization={Hopcroft Center on Computing Science, Huazhong University of Science and Technology},
    city={Wuhan},
    postcode={430074}, 
    country={China}}

\cortext[cor1]{The first two authors contribute equally}
\cortext[cor2]{Corresponding author}



\begin{abstract}
Meta-structures are widely used to define which subset of neighbors to aggregate information in heterogeneous information networks (HINs). In this work, we investigate existing meta-structures, including meta-path and meta-graph, and observe that they are initially designed manually with fixed patterns and hence are insufficient to encode various rich semantic information on diverse HINs. Through reflection on their limitation, we define a new concept called meta-multigraph as a more expressive and flexible generalization of meta-graph, and propose a stable differentiable search method to automatically optimize the meta-multigraph for specific HINs and tasks. As the flexibility of meta-multigraphs may propagate 
redundant messages, we further introduce a complex-to-concise (\meta) meta-multigraph that propagates messages from complex to concise along the depth of meta-multigraph. Moreover, we observe that the differentiable search typically suffers from unstable search and a significant gap between the meta-structures in search and evaluation. To this end, we propose a progressive search algorithm by implicitly narrowing the search space to improve search stability and reduce inconsistency. 
Extensive experiments are conducted on six medium-scale benchmark datasets and one large-scale benchmark dataset over two representative tasks, \ie, node classification and recommendation. Empirical results demonstrate that our search methods can automatically find expressive meta-multigraphs and \meta~meta-multigraphs, enabling our model to outperform state-of-the-art heterogeneous graph neural networks. 
\end{abstract}



\maketitle

\section{Introduction}
\label{sec:intro}
Heterogeneous information networks (HINs) consist of multiple types of nodes and/or edges. 
When simulating real-world systems, HINs can represent a wide range of entities and complex relationships, which is extremely useful in applications. For example, the citation network called ACM contains multiple node types, \ie, Author (A),  Paper (P), and Subject (S), as well as multiple edge types, such as Author$\xrightarrow{\rm writes}$Paper, Paper$\xrightarrow{\rm cites}$Paper, Paper$\xrightarrow{\rm belongs~to}$Subject.

To fully utilize the heterogeneity, various heterogeneous graph neural networks (HGNNs) have been proposed by combining graph neural networks (GNNs) with meta-structures~\cite{wang2019heterogeneous,han2020genetic,ji2021heterogeneous,DBLP:journals/kbs/ChangCHZZC22,DBLP:journals/kbs/Qian0L0022,DBLP:journals/kbs/XieZCZ21}, including meta-paths~\cite{Metapath,wang2019heterogeneous} and meta-graph~\cite{fang2016semantic,huang2016meta,zhao2017meta}. Within this framework, each node generates its representation by aggregating features of its neighbors defined by meta-structures. For example, the meta-path Author-Paper-Author (APA) represents the co-author relationship, which corresponds to multiple APA instances in the HIN.  
Then, one Author node can aggregate features from its meta-path-based neighbors (co-authors). One category of HGNNs employs hand-designed meta-structures to define neighbors, \eg,  HAN~\cite{wang2019heterogeneous} and MAGNN~\cite{fu2020magnn}. However, hand-designed meta-structures require rich domain knowledge that is extremely difficult to obtain in schema-rich or large-scale HINs. 
The second category of HGNNs implicitly learns meta-structures by attention to eliminate the need for hand-designed meta-structures, 
\eg, GTN~\cite{NIPS2019_9367} and HGT~\cite{hu2020heterogeneous}. However, these works typically suffer from high time or memory costs.  
More recently, in light of the achievement of neural architecture search (NAS) in convolutional neural networks (CNNs)~\cite{DBLP:conf/iclr/ZophL17, pham2018efficient,liu2018darts}, the third category of HGNNs employs NAS to automate the customization of meta-structures for specific datasets and tasks, including GEMS~\cite{han2020genetic} and DiffMG~\cite{DBLP:conf/kdd/DingYZZ21}. NAS-based HGNNs have accelerated the automation process in heterogeneous graph representation learning. 

However, there exists some weakness in these NAS methods for heterogeneous GNNs. 
GEMS~\cite{han2020genetic} uses an  evolutionary algorithm as the search strategy, making its search cost dozens of times as training a single GNN. 
Inspired by the simplicity and computational efficiency of differentiable architecture search~\cite{liu2018darts} in CNNs, DiffMG~\cite{DBLP:conf/kdd/DingYZZ21} proposes to search a meta-graph in a differentiable fashion, making the search cost on a par with training a single GNN once. Yet, it suffers from performance instability and often finds architectures worse than hand-designed models. This weakness motivates us to propose more stable search algorithms on HINs.



\begin{figure*}[!t]
    \centering
    
    \subfloat[Meta-path]{\includegraphics[width=0.45\linewidth]{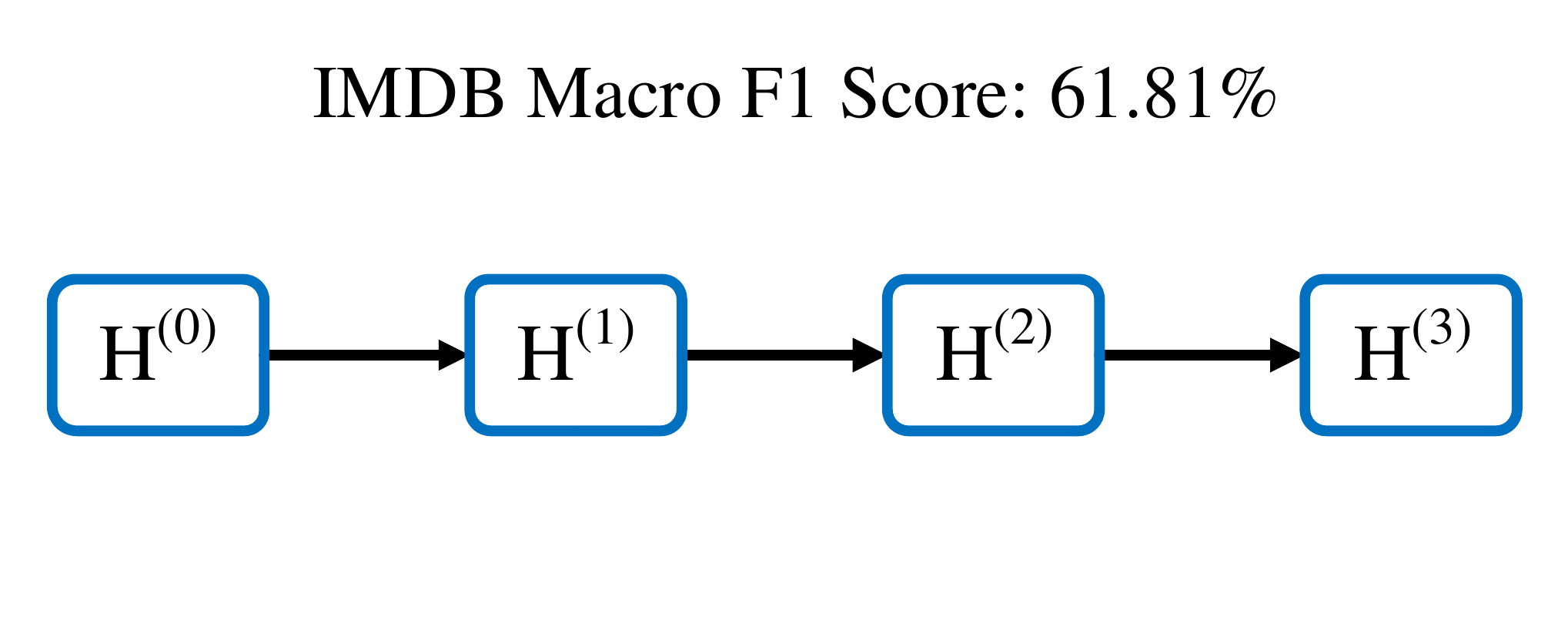}} 
    \hspace{1em}
    \subfloat[Meta-graph]{\includegraphics[width=0.45\linewidth]{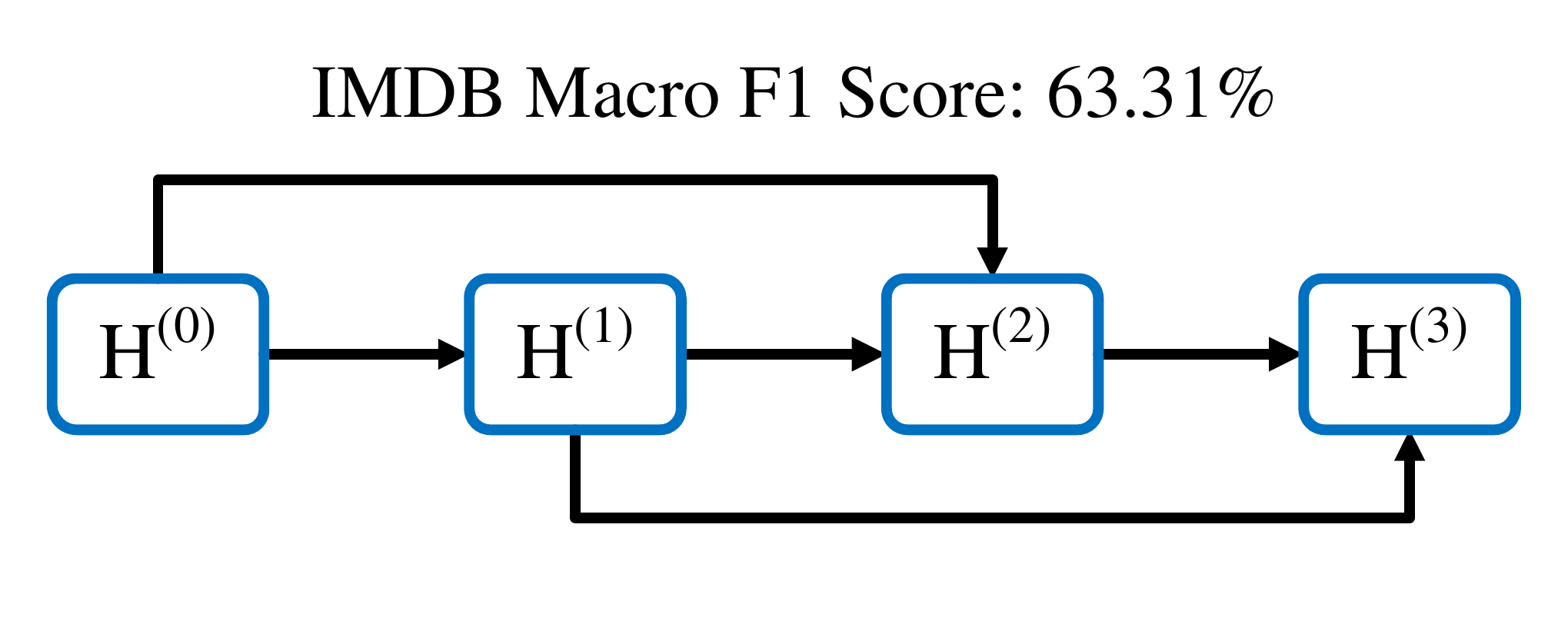}}\\
    \subfloat[Meta-multigraph]{\includegraphics[width=0.45\linewidth]{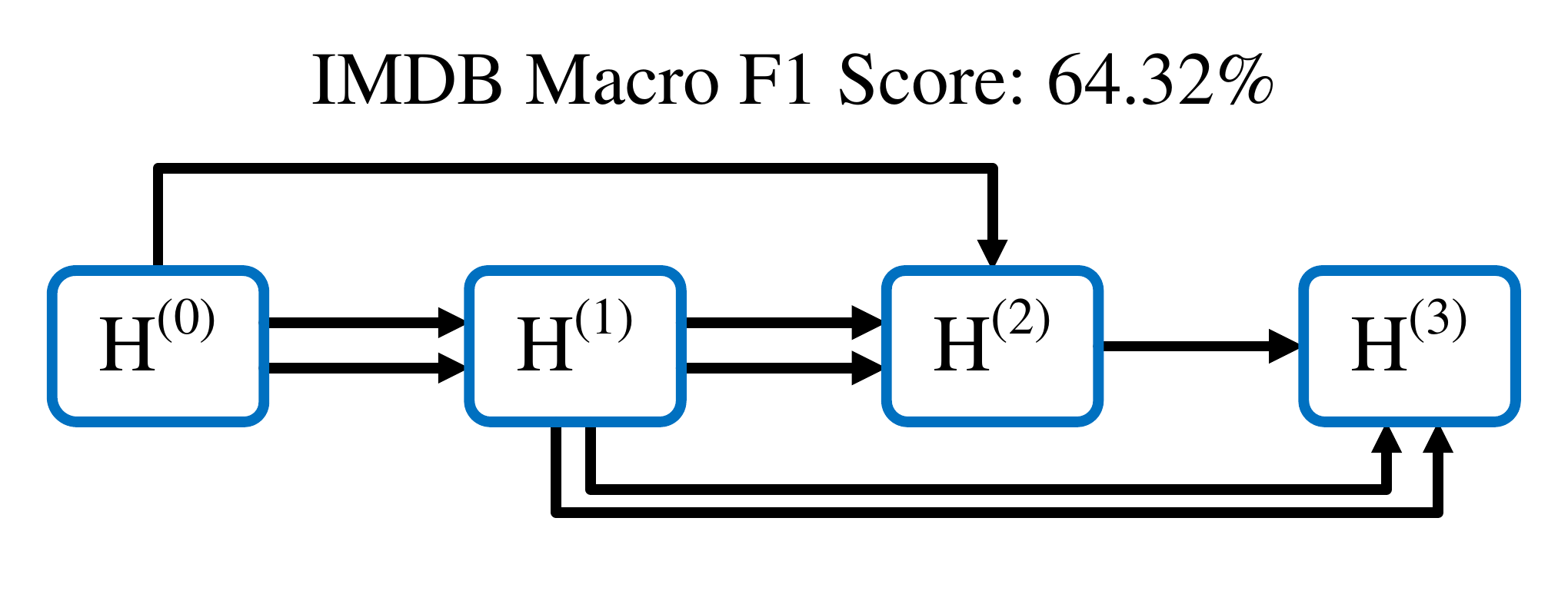}}
     \hspace{1em}
    \subfloat[\meta~meta-multigraph]{\includegraphics[width=0.45\linewidth]{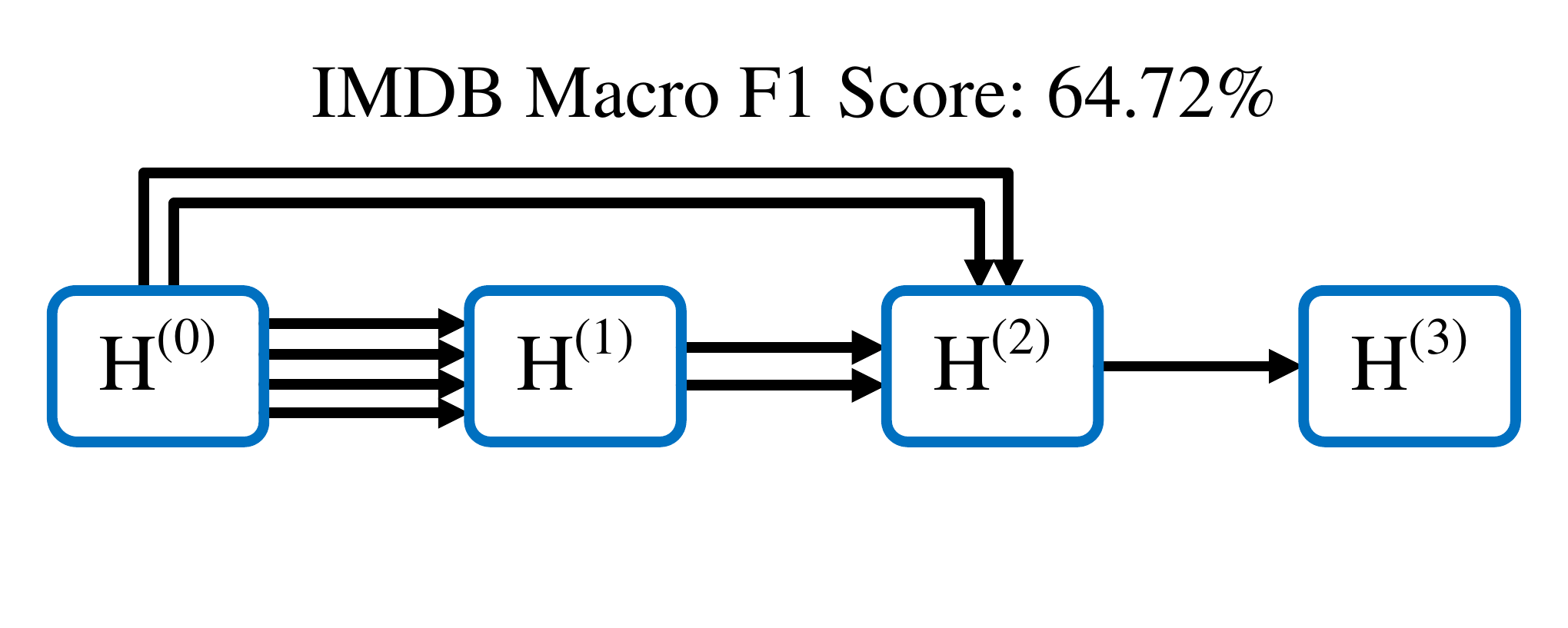}}
    \caption{Differences of (a) meta-path, (b) meta-graph, (c) meta-multigraph, and (d) complex-to-concise~meta-multigraph with depth $N=3$. 
    Our proposed meta-structures (c and d) are new for HINs, allowing propagating multiple  types of messages between hyper-nodes (see Definition~\ref{def:multigraph}). 
    Meta-multigraph is a natural generalization of meta-graph. Complex-to-concise~meta-multigraph is a variant of meta-multigraph that propagates messages from complex to concise. 
    }
\vspace{-1ex}
    \label{fig:compare}
\end{figure*}

Besides, existing works have a common 
issue of inflexibility in meta-structures. For HGNNs with hand-designed meta-structures, the meta-structures are predefined with fixed patterns, so they have limited capacity to encode various rich semantic information on diverse HINs and tasks. HGNNs employing attention or NAS can learn dataset-dependent meta-structures but only considers meta-paths or meta-graphs to guide the GNNs to propagate messages. However, meta-paths and meta-graphs are initially applied for hand-designed heterogeneous GNNs. They  
would restrict the searched meta-structures to inflexible topology, which further limits the performance of HGNNs.  
Therefore, 
it is still an open problem on 
whether better meta-structures exist for HGNNs (see Table~\ref{tab:hingnn} for a summary).
\begin{table}[t]
\small
	\caption{Classification of heterogeneous GNNs based on the utilized semantic information. Semantics indicate what kind of semantic information a method utilizes. }
	\label{tab:hingnn}
	\begin{threeparttable}[b]
    \footnotesize
    \resizebox{0.48\textwidth}{!}
    {
	\begin{tabular}{c|c} 
	\toprule
	\textbf{Semantics} & \textbf{Method}        \\
	\midrule
	\multirow{2}{*}{Meta-path} & HAN~\cite{wang2019heterogeneous}, GTN~\cite{NIPS2019_9367}, 
	MAGNN~\cite{fu2020magnn}, \\
	&HGT~\cite{hu2020heterogeneous}, 
	HGNAS~\cite{gao2021heterogeneous}\\
	\midrule
	Meta-graph & GEMS~\cite{han2020genetic}, 
	DiffMG~\cite{DBLP:conf/kdd/DingYZZ21} \\
    \midrule
	Meta-multigraph & \mone~(ours)\\
	\midrule
	\meta~meta-multigraph & \model~(ours)\\
	\bottomrule
	\end{tabular}
 }
	\end{threeparttable}
\end{table}


This paper is an extended and improved version of our conference paper~\cite{DBLP:journals/corr/abs-2211-14752}, in which we propose the Partial Message Meta-Multigraph search (PMMM), consisting of a more flexible and expressive meta-structure called meta-multigraph, and a partial message search algorithm to overcome the instability for differentiable search on HINs. In this paper, we make a more comprehensive exploration of different meta-structures and observe that the performance of HGNNs improves along with the complexity of the searched meta-structures (see Fig.~\ref{fig:compare}). We further propose a new algorithm called the Progressive Complex-to-concise Meta-Multigraph search (\model), containing a more expressive complex-to-concise (\meta) meta-multigraph, and a progressive search algorithm to overcome both the inconsistency and instability in existing NAS-based HGNNs.

Specifically, to stabilize the differentiable search, PMMM randomly selects partial candidate message-passing types to update per iteration to ensure all candidate paths be equally and fully trained and decouple the joint optimization of paths. To derive flexible topology, PMMM searches for a novel meta-multigraph by selecting the top-$k$ most promising candidate message-passing types for aggregation. Searching for a meta-multigraph is a free performance-enhancing strategy, as it can encode more flexible and sophisticated semantic information in HINs compared with a meta-path or meta-graph. 

To bridge the gap between the meta-structures in the search and evaluation stages of the differentiable framework~\cite{yang2020ista,yang2021towards}, \model~employs a progressive search algorithm to implicitly narrow the search space by progressively reducing the number of activated paths. Then, the collection of the activated paths at the end of the search directly makes up our target-net. Consequently, a high-performance super-net corresponds to an effective target-net. Due to the progressive strategy, the whole search stage is stable.  
Additionally, because meta-multigraphs' flexibility may result in the propagation of redundant messages, 
\model~searches for a more expressive \meta~meta-multigraph by selecting the top-$n$ instead of the top-1 most promising candidate message-passing types related to the depth of meta-multigraph. 
Experiments on seven widely used and diverse datasets demonstrate that \mone~and \model~outperform state-of-the-art HGNNs on node classification and recommendation tasks. 

Our main contributions are summarized as follows: 

\begin{itemize}
\item  We propose a novel meta-structure, termed meta-multigraph, 
that is more expressive and flexible than existing meta-structures. 

\item We propose the first stable differentiable search algorithm on HINs, called PMMM, which can consistently discover promising meta-multigraphs that outperform hand-designed meta-structures. 

\item We further propose \model, consisting of a more expressive complex-to-concise meta-multigraph to avoid propagating redundant messages, and a progressive differentiable search algorithm to bridge the gap between the meta-structures in search and evaluation.


\item Thorough experiments demonstrate that our \mone~and \model~can find effective meta-structures and achieve state-of-the-art performance on medium-scale and large-scale datasets. 
\end{itemize}

The rest of this paper is organized as follows.  Section~\ref{sec:Definitions} introduces necessary definitions, including meta-path, meta-graph, and the new conception, meta-multigraph and C2C meta-multigraph. Secion~\ref{sec:R} presents the related works of heterogeneous graph neural networks and heterogeneous graph neural architecture search. 
Section~\ref{sec:method} describes the details of our proposed PMMM and PCMM algorithms, and how they improve stability and consistency, respectively. Section~\ref{sec:experiments} presents experimental results and analyses. Section~\ref{sec.Conclusion} contains the concluding remarks. 


\begin{figure*}[!t]
    \centering
    \subfloat[An academic network with network schema]{\includegraphics[width=0.42\linewidth]{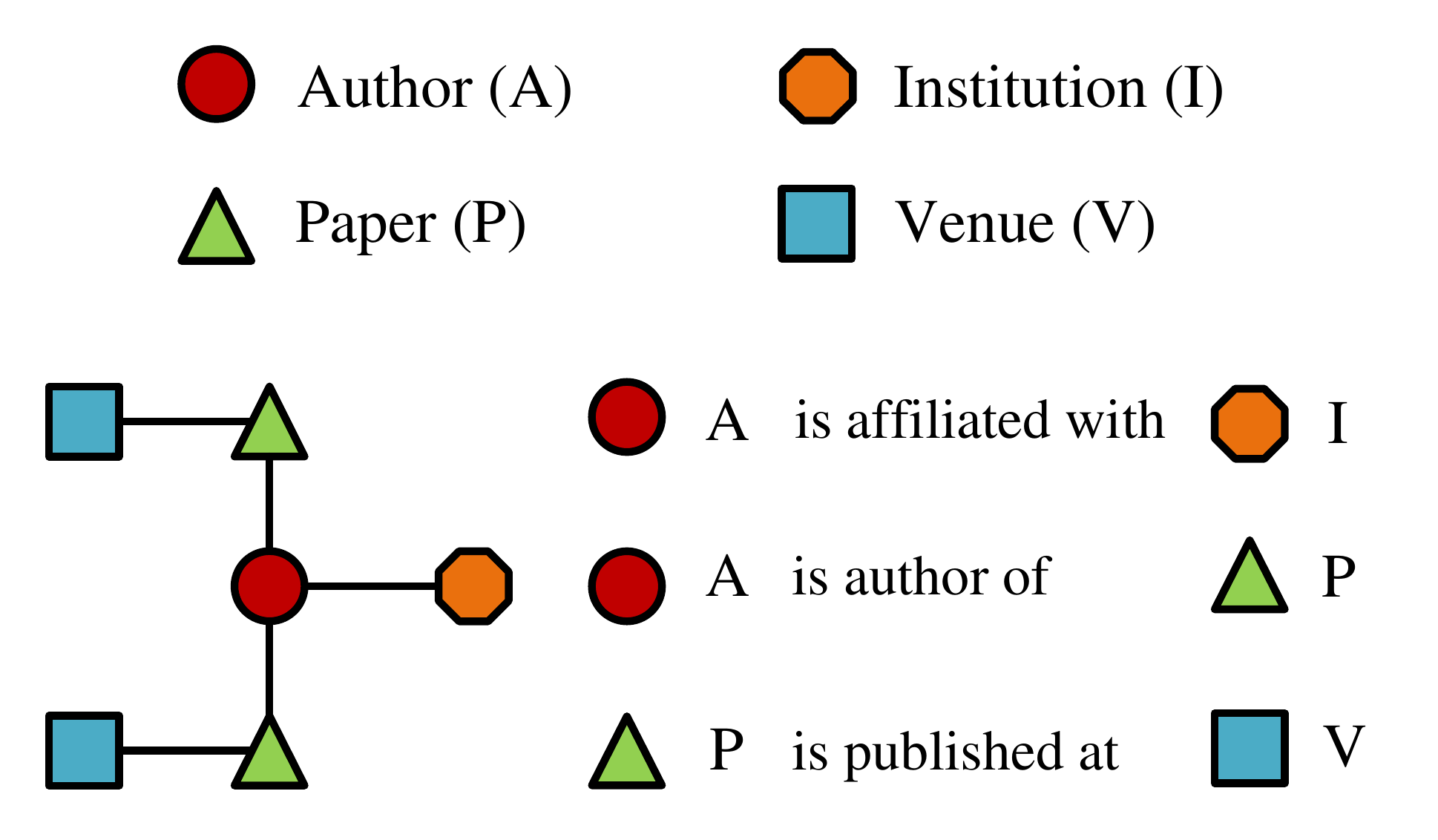}}
    \hspace{1em}
    \subfloat[Meta-path]{\includegraphics[width=0.42\linewidth]{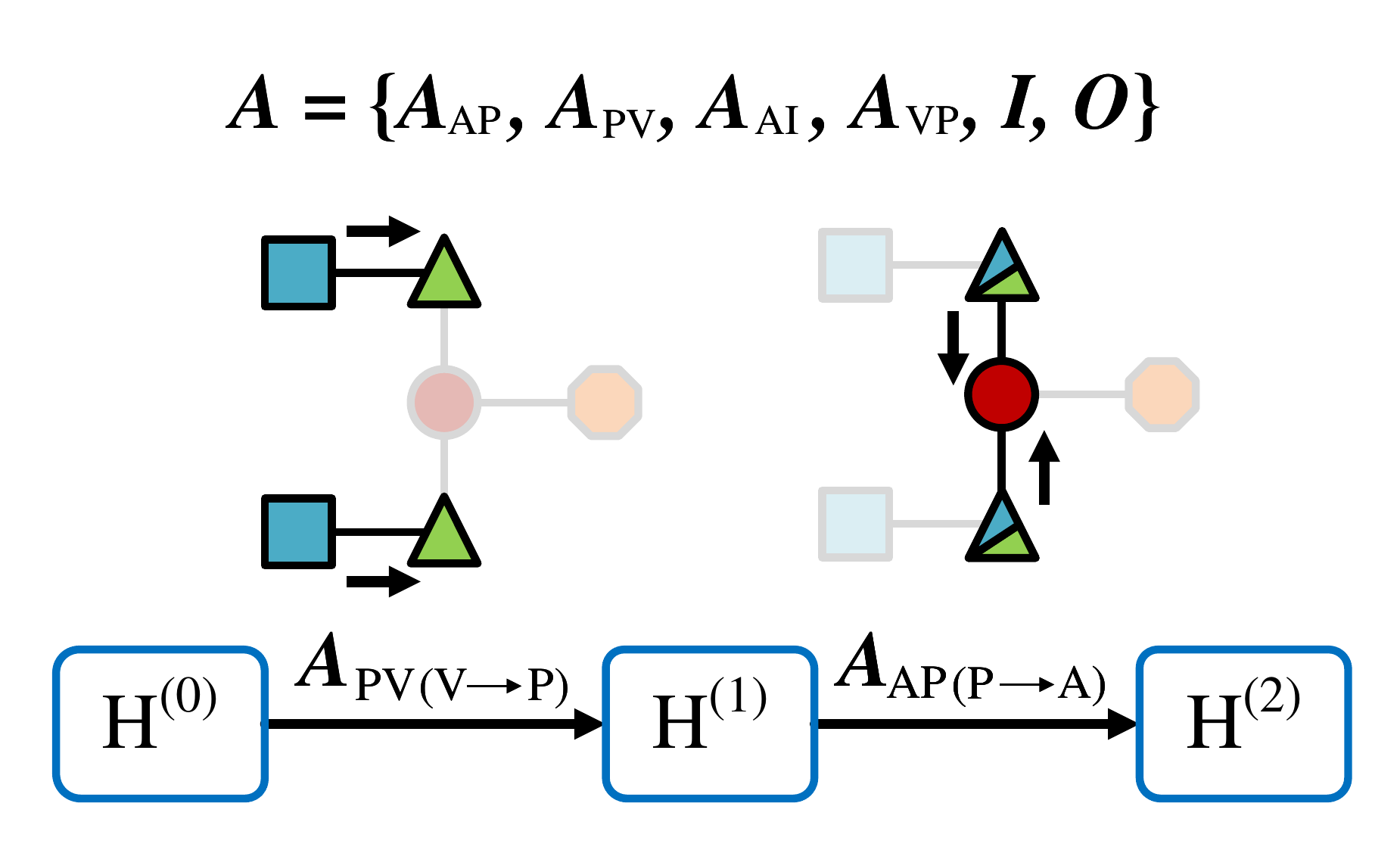}} \\
    \subfloat[Meta graph]{\includegraphics[width=0.42\linewidth]{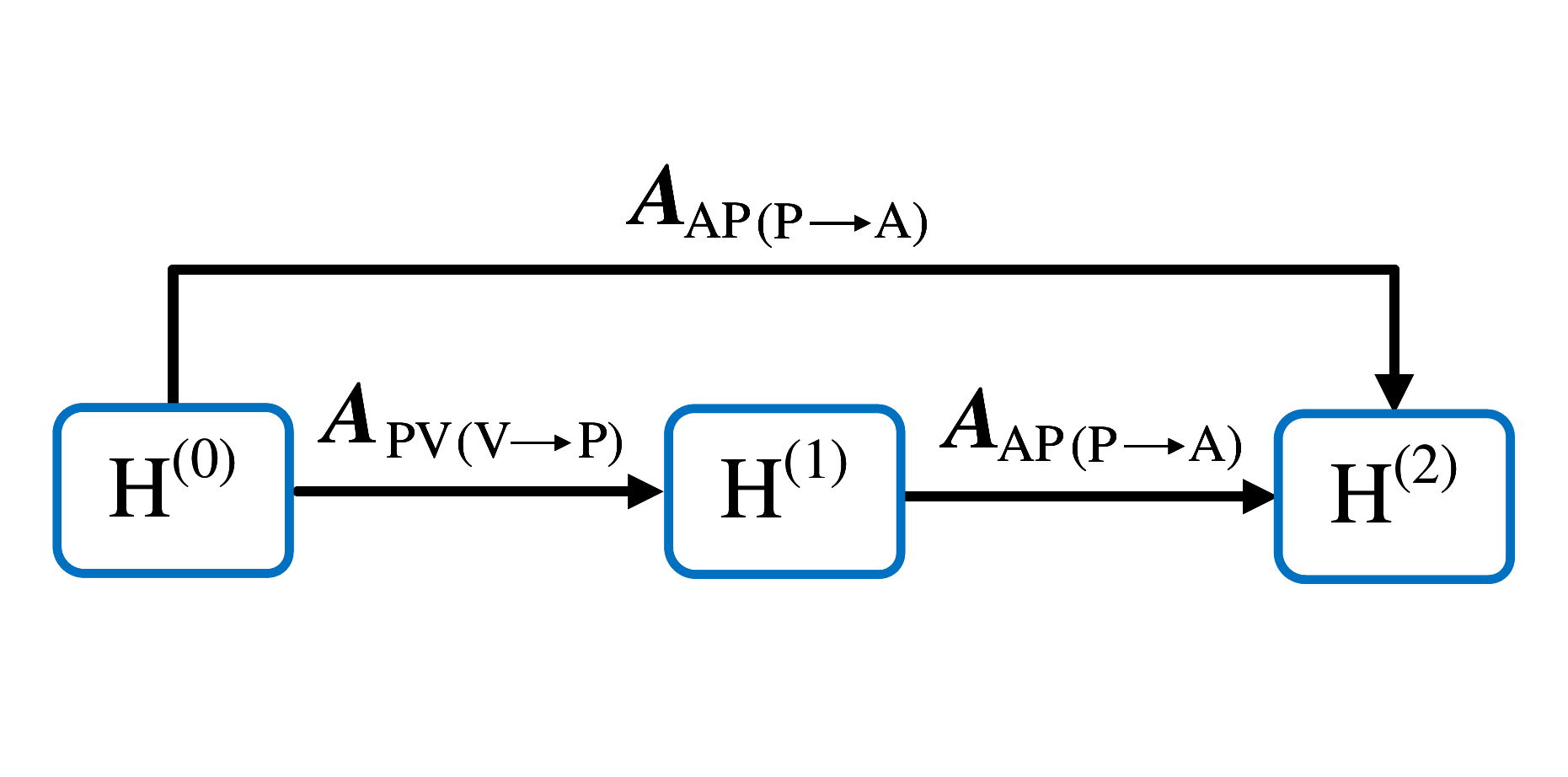}}
    \hspace{1em}
    \subfloat[Meta  multigraph]{\includegraphics[width=0.42\linewidth]{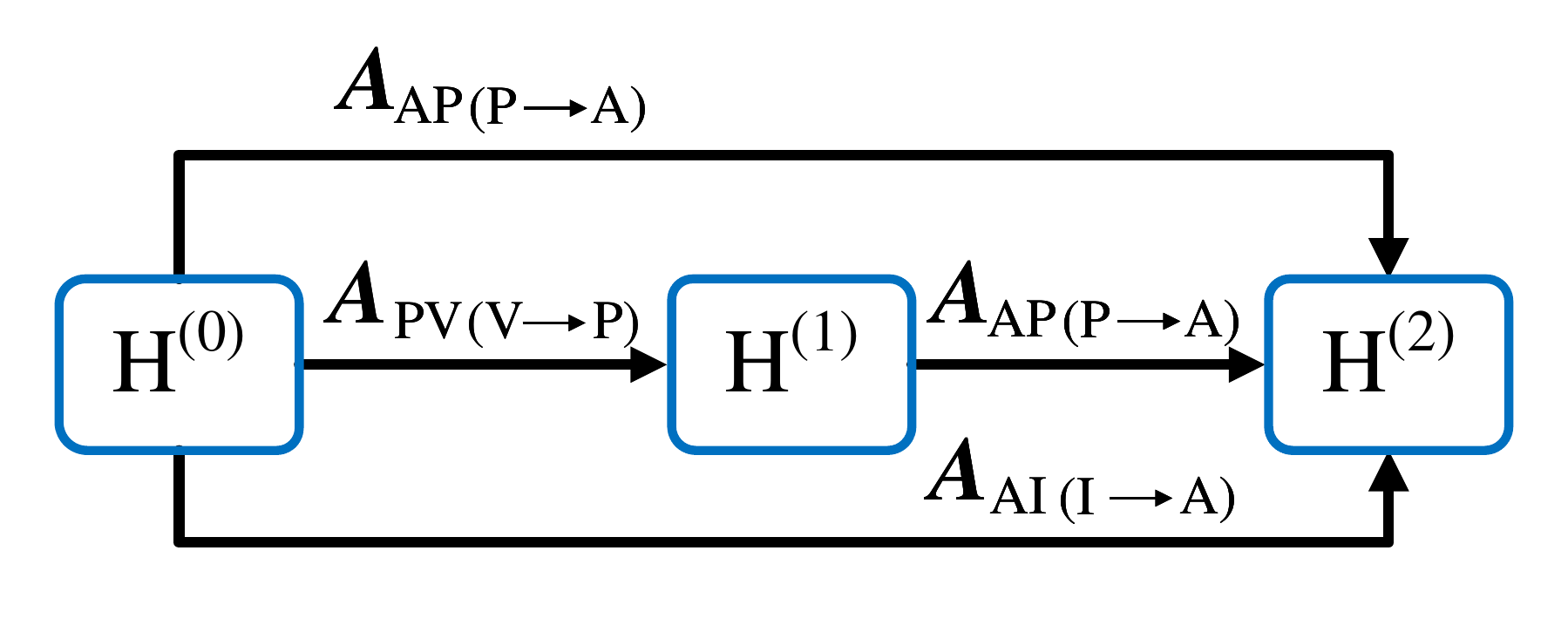}}

    \caption{Illustration of (a) an example academic network, (b) meta-path, (c) meta-graph, and (d) meta-multigraph with depth $N=2$. $\bm{A}$ is the set of candidate edge types between two hyper-nodes (see Definition~\ref{def:multigraph}). 
    A meta-multigraph allows propagating multiple types of messages between hyper-nodes, which cannot be realized by a meta-path or a meta-graph.
    }
    \label{fig:multigraph}
\end{figure*}

\section{Definitions}
\label{sec:Definitions}
In this section, we first provide the existing definitions on Heterogeneous Information Network(HIN)~\cite{sun2011pathsim},
meta-path~\cite{Metapath} and meta-graph~\cite{zhao2017meta}, then we introduce the new conception called
meta-multigraph.

\subsection{Heterogeneous Information Network}
\begin{definition}[Heterogeneous Information Network~\cite{sun2011pathsim}]
A heterogeneous information network (HIN) is defined as a directed graph $\mathcal{G}=\{\mathcal{V}, \mathcal{E}, \mathcal{T}, \mathcal{R}, f_{\mathcal{T}}, f_{\mathcal{R}}\}$, where $\mathcal{V}$ signifies the set of nodes and $\mathcal{E}$ denotes the collection of edges, $\mathcal{T}$ is the node-type set and $\mathcal{R}$ is the edge-type set. Each node $v \in \mathcal{V}$ is associated with type mapping function $f_{\mathcal{T}}(v) \in \mathcal{T}$ and each edge $e \in \mathcal{E}$ is associated with type mapping function $f_{\mathcal{R}}(e) \in \mathcal{R}$. 
We define its network schema as 
$\mathcal{S}=\{\mathcal{T}, \mathcal{R}\}$ where $\lvert \mathcal{T} \rvert >1$ or $\lvert \mathcal{R} \rvert >1$. 
\end{definition}

For $r \in \mathcal{R}$, we use $\bm{\mathcal{A}}_{r}$ to denote the adjacency matrix formed by the edges of type $r$ in $\mathcal{G}$ and $\bm{\mathcal{A}}$ to denote the collection of all $\bm{\mathcal{A}}_{r}$'s. For example, in Figure~\ref{fig:multigraph} (b), $\bm{\mathcal{A}}_{AP}$ represents directed edges from paper nodes (P) to author nodes (A). 




\subsection{Meta-path and meta-graph}

\begin{definition}[Meta-path~\cite{Metapath}]
\label{def:path}
A meta-path is a path with length $l$ defined on the schema $\mathcal{S}=\{\mathcal{T}, \mathcal{R}\}$ of $\mathcal{G}$, and is denoted in the form of $t_1\stackrel{r_1}{\longrightarrow}t_2\stackrel{r_2}{\longrightarrow}...\stackrel{r_l}{\longrightarrow}t_{l+1}$, where $t_1,...,t_{l+1} \in \mathcal{T}$ and $r_1,...,r_l \in \mathcal{R}$. In the underlying HIN, one meta-path can correspond to multiple meta-path instances in $\mathcal{G}$.
\end{definition}


Meta-graph~\cite{zhao2017meta} allows the in-degree of each node type (except the source node type) to be greater than 1. As compared to meta-paths, meta-graphs can represent 
more complex semantic relations. 

\begin{definition}[Meta-graph]
\label{def:graph}
A meta-graph is a directed acyclic graph on the network schema $\mathcal{S}$ with a single source node $t_s \in \mathcal{T}$ (with zero in-degree), a single target node 
$t_e \in \mathcal{T}$ (with zero out-degree), and the connected internal nodes. 
\end{definition}

\vspace{-1ex}

\subsection{Meta-multigraph}

A meta-graph can only propagate one message-passing type between two nodes, which is insufficient to encode the rich semantic information on HINs. 
An example is shown in Fig.~\ref{fig:multigraph}. The meta-multigraph in (d) allows the author to aggregate the initial information of both paper and institution, while the meta-graph in (c) can only aggregate the initial information of either paper or institution. Based on the above 
observations, we define meta-multigraph to facilitate the description of our method.

\begin{figure*}[tb]
    \centering
    \includegraphics[width=0.90\linewidth]{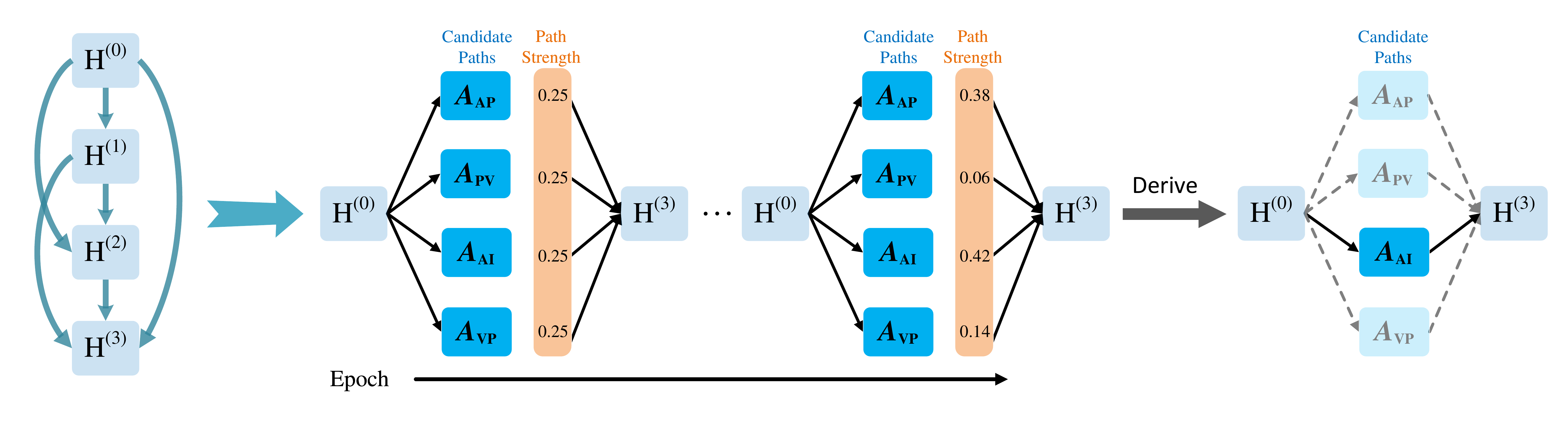}
    \caption{A diagram of the basic framework of differentiable meta-multigraph search. We only show the search process between $\bm{H}^{(0)}$ and $\bm{H}^{(3)}$ for ease of illustration. The target-net is derived by keeping the strongest path between two hyper-nodes and has a significant gap with the super-net in search.
    }
    \vspace{-1ex}
    \label{fig:darts}
\end{figure*}

\begin{definition}[Meta-multigraph]
\label{def:multigraph}
A meta-multigraph defined on network schema $\mathcal{S}$ is a directed acyclic multigraph consisting of multi-edges and hyper-nodes. Each multi-edge $\mathcal{R}_i \subseteq \mathcal{R}$ is a set of multiple edges, and each hyper-node $\mathcal{T}_j \subseteq \mathcal{T}$ is the set of heads of its incoming multi-edges and the tails of its outgoing multi-edges. A meta-multigraph contains a single source hyper-node, a single sink (target) hyper-node, and the connected internal hyper-nodes. Each node in a hyper-node can correspond to multiple node instances and each edge in a multi-edge can correspond to multiple edge instances in graph $\mathcal{G}$.
\end{definition}

A complex-to-concise meta-multigraph is a variant of the meta-multigraph. The number of edge types in its multi-edges decreases with the depth, indicating that the message propagation is performed from 
complex to concise. 

To integrate meta-path and meta-graph into the automatic framework, we expand the node type to hyper-node in Definitions~\ref{def:path} and~\ref{def:graph}, generalizing the meta-path and meta-graph to  more flexible definitions on the network schema.

\section{Related Works}
\label{sec:R}
\textbf{Heterogeneous Graph Neural Networks (HGNNs)} have been successfully applied to model HINs due to their ability to utilize rich and diverse information. To effectively guide the information propagation in HINs, most HGNNs define neighbors by meta-structures, including meta-paths~\cite{Metapath} and meta-graphs~\cite{fang2016semantic, huang2016meta, zhao2017meta}. 
One category of HGNNs employs hand-designed meta-structures to define neighbors, \eg,  HAN~\cite{wang2019heterogeneous} and MAGNN~\cite{fu2020magnn}. The second category of HGNNs implicitly learns meta-structures by attention to eliminate the need for hand-designed meta-structures, \eg, GTN~\cite{NIPS2019_9367} and HGT~\cite{hu2020heterogeneous}. Compared to these methods, our PMMM and \model~can find efficient meta-structures by filtering out unrelated edge types. 



\textbf{Heterogeneous Graph Neural Architecture Search (HGNAS)} is proposed to automatically find the best HGNNs for the given specific graph task. Recently, in light of the achievement of neural architecture search (NAS) in convolutional neural networks (CNNs), numerous NAS-based works have been proposed to obtain data-specific homogeneous GNNs for specific datasets and tasks~\cite{zhou2019auto,qin2021graph,zhao2021search,wei2022designing,zhang2022pasca}. 
Meanwhile, only a few works 
attempt to employ NAS in HINs due to the complexity of  semantic relationships. 
GEMS~\cite{han2020genetic} employs an evolutionary algorithm to search for meta-graphs for recommendation. To overcome the inefficiency of the evolutionary algorithm, DiffMG~\cite{DBLP:conf/kdd/DingYZZ21} employs a differentiable algorithm to search for meta-graphs. 
In contrast, 
our PMMM and \model~can perform an efficient and stable search for different tasks on HINs. Furthermore, our model searches for a meta-multigraph, which shows better diversity and a stronger capacity to capture complex semantic information.

\section{Methodology }
\label{sec:method}
In this section, we first briefly show the framework of the proposed differentiable meta-multigraph search on HINs. Then, we propose a partial message search algorithm for improving stability and introduce the meta-multigraph. Next, we develop our proposed progressive search algorithm, showing how it works to improve consistency and stability. Finally, we present how to generate different meta-structures for exploring the relationship between the performance of HGNNs and the types of meta-structures.

\subsection{Differentiable Meta-multigraph Search}
\label{m1}

Differentiable meta-multigraph search is developed on differentiable meta-graph search  (DiffMG)~\cite{DBLP:conf/kdd/DingYZZ21}. The differentiable meta-multigraph search aims to find proper message-passing types between different hyper-nodes automatically. 
The framework of differentiable meta-multigraph search consists of two stages. 
At the search stage, we train a super-net, from which sub-networks can be sampled exponentially. The super-net is a 
directed acyclic multigraphs. Each multi-edge in the directed acyclic multigraph is the set of candidate message-passing types, \ie, candidate paths. 
Each message-passing type is associated with an aggregator to perform a feature aggregation. 
As shown in Fig.~\ref{fig:darts}, each candidate path is weighted by the path strength calculated by architecture parameters, which are jointly optimized with the super-net weights in an end-to-end manner. The goal of the search stage is to determine the architecture parameters. 
At the evaluation stage, the strongest sub-network is preserved as the target-net by pruning redundant paths based on the searched architecture parameters. The target-net is then retrained from scratch to gain the final results.

Recall that for a homogeneous network, the GCN layer can be represented as:
\begin{equation}
\label{eq:gcn}
\bm{X}^{(l+1)} = \sigma (\hat{\bm{A}}\bm{X}^{(l)}\bm{\Theta}^{(l)}),
\end{equation}
where $\bm{X}^{(l)} \in \mathbb{R}^{N \times d}$ and $\bm{X}^{(l+1)} \in \mathbb{R}^{N \times d}$ denote the input
and output node representations of layer $l$, respectively. $\bm{\Theta}$ denotes a shared hidden weight matrix, and $\sigma$ denotes the activation function. The normalized adjacency matrix $\hat{\bm{A}}$ is used to represent information propagation between one-hop neighbors. However, for heterogeneous networks, \eqref{eq:gcn} cannot capture the differences among various node types, leading to a loss of semantic information.

To preserve semantic information, it is crucial to distinguish messages that propagate along different edge types. Furthermore, distinct combinations of messages from different edge types may generate diverse node representations. 
To depict the message-passing process, we extend Eq.~\ref{eq:gcn} to a heterogeneous message-passing layer by selecting edge types and establishing suitable connections among them:
\begin{equation}
\label{eq:layer}
    \bm{X}^{(l+1)} = \sigma \left( F_{\bm{\mathcal{A}}} \left( \bm{\hat{X}}^{(l)}\bm{\Theta} \right) \right ),
\end{equation}
where $\bm{\hat{X}}^{(l)}$ represents the output of the projected feature by type-specific transformation~\cite{wang2019heterogeneous, fu2020magnn} which maps features of different node types into a common latent space. The $F_{\bm{\mathcal{A}}}(\cdot)$ characterizes the message-passing process that is aware of the edge types, \ie, $\bm{\mathcal{A}}$.

We define the search space of $F_{\bm{\mathcal{A}}}(\cdot)$ as a directed acyclic multigraph, in which the hyper-node set $\bm{H}=\{\bm{H}^{(0)},\bm{H}^{(1)},\cdots,\bm{H}^{(n)},\cdots,\bm{H}^{(N)}\}$ denotes all the hyper-nodes in the message-passing process and the multi-edge set $E= \{\mathcal{R}^{(i,j)}|0 \le i<j\le N\}$ contains all the multi-edges between hyper-nodes. Each hyper-node $\bm{H}^{(n)} \subseteq \mathcal{T}$. $\bm{H}^{(0)}$ and $\bm{H}^{(N)}$ is the input and output of the meta-multigraph, respectively. 
Each multi-edge $\mathcal{R}^{(i,j)} \subseteq \mathcal{R}$ contains multiple paths, corresponding to candidate message-passing types.  
The key idea of differentiable meta-multigraph search is to formulate the information propagated from $\bm{H}^{(i)}$ to $\bm{H}^{(j)}$ as a weighted sum over all candidate paths, namely: 

\begin{align}
\bm{H}^{(j)}=\sum_{i<j}\sum_{r \in \mathcal{R}^{(i,j)}} p^{(i,j)}_r {f}\left(\bm{\mathcal{A}}^{(i,j)}_r,\bm{H}^{(i)}\right),
\label{eq:forward}
\end{align}
\begin{align}
p^{(i,j)}_r=\exp(\alpha^{(i,j)}_r)/\sum_{r \in \mathcal{R}^{(i,j)}}\exp(\alpha^{(i,j)}_r).
\label{eq:softmax}
\end{align}

Here ${f}(\bm{\mathcal{A}}^{(i,j)}_r,\bm{H}^{(i)})$ denotes one message-passing step that aggregates
$\bm{H}^{(i)}$ along message-passing type $\bm{\mathcal{A}}^{(i,j)}_r$. Following DiffMG~\cite{DBLP:conf/kdd/DingYZZ21}, ${f}(\cdot,\cdot)$ is a mean aggregator. 
$\alpha^{(i,j)}_r$ indicates the architecture parameters of 
$\bm{\mathcal{A}}^{(i,j)}_r$, and $p^{(i,j)}_r \in (0,1]$ denotes the corresponding path strength calculated by a softmax over $\alpha^{(i,j)}_r$. ${f}(\cdot, \cdot)$ can be any aggregation function. Following DiffMG~\cite{DBLP:conf/kdd/DingYZZ21}, we employ a simple mean aggregator.

The parameter update in Eq.~\ref{eq:forward} involves a bilevel optimization problem~\cite{anandalingam1992hierarchical, colson2007overview,xue2021rethinking}: 
\begin{align}
	\min_{\bm{\alpha}} \quad & \mathcal{L}_{val}(\bm{\omega}^*(\bm{\alpha}), \bm{\alpha}) \label{eq:outer} \\
	\text{s.t.} \quad &\bm{\omega}^*(\bm{\alpha}) = \mathrm{argmin}_{\bm{\omega}} \enskip \mathcal{L}_{train}(\bm{\omega}, \bm{\alpha}) ,
	\label{eq:inner}
\end{align}
where $\mathcal{L}_{train}$ and $\mathcal{L}_{val}$ denote the training and validation loss, respectively. The goal of the search stage is to find $\bm{\alpha}^*$ that minimizes $\mathcal{L}_{val}(\bm{\omega}^*, \bm{\alpha}^*)$. 

At the evaluation stage, we derive a compact meta-multigraph by pruning redundant paths based on path strengths $p^{(i,j)}_r$ determined by the architecture parameters. Each edge in the meta-multigraph is associated with ${f}(\cdot,\cdot)$ to represent a message-passing step. The final HGNN architecture is made up of message-passing steps and the module related to downstream tasks, such as MLP. The meta-multigraph-based HGNN is then retrained from scratch to generate node representations for different downstream tasks, \ie, node classification and recommendation (link prediction).

\subsection{Partial Message Meta-multigraph Search for stability}
\label{m2} 

DiffMG is the most related work to our differentiable meta-multigraph search. Although DiffMG is efficient and outperforms existing baselines, its limitations lie in its instability. In each iteration, DiffMG samples one candidate message-passing type on each edge for forward propagation and backpropagation, and the strongest message-passing type is most likely to be sampled, resulting in random and insufficient training.  
As illustrated in Fig.~\ref{fig:seed} of the experiments, DiffMG is only effective in a few random seeds and the performance dramatically declines in most random seeds. 

Intuitively, to address the instability issue in DiffMG, we wish to ensure that all message-passing types are equally and fully searched in the search stage. An alternative solution is to employ Eq.~\ref{eq:forward}, which trains all possible message-passing steps together and formulates the information propagated as a weighted sum over all the paths. However, the architecture parameters of various paths in Eq.~\ref{eq:forward} are deeply coupled and jointly optimized. 
The greedy nature of the differentiable methods inevitably misleads the architecture search due to the deep coupling~\cite{guo2020single}, especially when the number of candidate paths  
is large. It motivates us to propose a new partial message meta-multigraph (PMMM) search algorithm to achieve a stable meta-multigraph search as well as overcome the coupling in optimization.

\begin{figure*}[tb]
    \centering
    \includegraphics[width=0.85\linewidth]{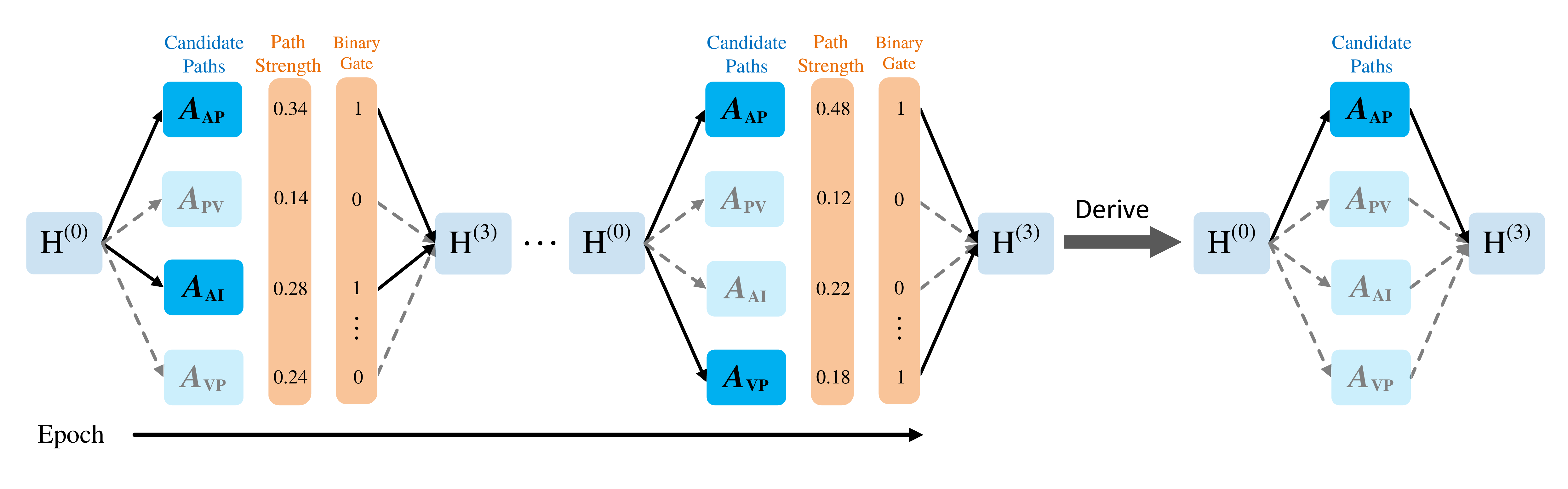}
    \vspace{-1ex}
    \caption{A diagram of the search algorithm of PMMM. The binary gate overcomes the coupling optimization issue. We only derive the strongest path for ease of illustration. 
    }
    \vspace{-0.5ex}
    \label{fig:partial}
\end{figure*}

To overcome the coupling optimization, as illustrated in Fig.~\ref{fig:partial}, 
we define a binary gate $M^{(i,j)}_r$ for each message-passing type,  
which assigns $1$ to the selected message-passing types and $0$ to the masked ones. 
Specifically, we let paths in each multi-edge be sampled equally and independently, and we set the proportion of $M^{(i,j)}_r=1$ to $1/p$ by regarding $p$ as a hyper-parameter.  
Then we can get the set of all active paths between hyper-node $\bm{H}^{(i)}$ and $\bm{H}^{(j)}$: 
\begin{equation}
S^{(i,j)}=\{r|M^{(i,j)}_r=1,\forall r \in \mathcal{R}^{(i,j)} \}.
\label{eq:sample}
\end{equation}


By introducing the binary gates, only $1/p$ paths of message-passing steps are activated. 
Then we formulate the information propagated from $\bm{H}^{(i)}$ to $\bm{H}^{(j)}$ as a weighted sum over activated candidate message-passing steps: 
\begin{align}
\bm{H}^{(j)}=\sum_{i<j}
\sum_{r\in S^{(i,j)}}p^{(i,j)}_r {f}\left(\bm{\mathcal{A}}^{(i,j)}_r,\bm{H}^{(i)}\right),
\label{eq:forward_ours1}
\end{align}
where path strength $p^{(i,j)}_r$ is calculated by Eq.~\ref{eq:softmax}.
Since message sampling masks $M^{(i,j)}$ are involved in the computation graph, parameters updated in Eq.~\ref{eq:forward_ours1} can be calculated through backpropagation. 

When updating the parameters, we need to address the bilevel optimization problem in Eq.~\ref{eq:outer} and Eq.~\ref{eq:inner}. Here, we transform the first-order approximation~\cite{DBLP:conf/iclr/LiuSY19} into our optimization, which empirically approximates $\bm{\omega}^*(\bm{\alpha})$ with the $\bm{\omega}$ dynamically maintained during training. Specifically, the updates of architecture parameters and network weights are performed alternately. 
The overall algorithm is presented in Algorithm 1.
\begin{algorithm}[!t]
	\caption{Search algorithm}
	\label{alg:ACA}
	\begin{algorithmic}[1]
		\REQUIRE~~\\
		Network weights $\bm{\omega}$; Architecture parameters $\bm{\alpha}$;  \\
		Number of iterations $T$; Sampling proportion $1/p$.\\
        \ENSURE ~~\\
        A meta-multigraph.
        \STATE Initialize network weights $\bm{\omega}$ and architecture parameters $\bm{\alpha}$
 		\FOR{each iteration $t \in [1,T]$}
\STATE Randomly sample $1/p$ candidate message passing steps in each multi-edge. 
The collection of network weights and architecture parameters $\bm{\alpha}^{(i,j)}_k$ of 
sampled paths 
is denoted as $\bar{\bm{\omega}}$ and $\bar{\bm{\alpha}}$, respectively
\STATE Update weights $\bar{\bm{\omega}}$ by  $\nabla_{\bar{\bm{\omega}}}{\cal L}_{train}(\bar{\bm{\omega}}, \bar{\bm{\alpha}})$
\STATE Execute step $3$ again
\STATE Update parameters $\bar{\bm{\alpha}}$ by  $\nabla_{\bar{\bm{\alpha}}}{\cal L}_{val}(\bar{\bm{\omega}}, \bar{\bm{\alpha}})$
\ENDFOR
\STATE Derive the meta-multigraph based on Eq.~\ref{eq:softmax},~\ref{eq:tau1}, and~\ref{handcrafted} 
\RETURN The derived meta-multigraph
	\end{algorithmic}
	
\end{algorithm}

	

After we have completed training the architectural parameters, we can derive the compact meta-structures by pruning redundant paths. To our knowledge, all existing differentiable search algorithms for CNNs choose a path with the highest strength on each edge because retaining multiple paths means employing multiple types of operations between two node representations, which degrades the performance in most cases.

DiffMG inherits the derivation strategy of these methods to generate meta-graphs. However, instead of operations, DiffMG searches the message-passing types that determine which messages are propagated between different types of node representations in a meta-graph. Considering that an HIN consists of multiple node types and edge types, deriving a single message-passing type between two different node representations is insufficient and inflexible to encode the rich semantic information. As shown in Fig.~\ref{fig:multigraph} (d), there are multiple message-passing types between $\bm{H}^{(0)}$ and $\bm{H}^{(2)}$, which cannot be learned by traditional derivation strategy. Another extreme example is when all candidate message-passing types are necessary for two hyper-nodes, retaining one message-passing type will have extremely low performance.  
Another issue caused by deriving a single path is that some effective message-passing types with similar but weaker path strengths will be dropped. For example, in Fig.~\ref{fig:darts}, $\bm{\mathcal{A}}_{AP}$ and $\bm{\mathcal{A}}_{AI}$ have similar path strength, $\bm{\mathcal{A}}_{AP}$ will be dropped if we only derive the strongest path.  
So simply selecting the message-passing type with the highest path strength may reject potentially good meta-structures. 

To address the above issues, we propose to derive a meta-multigraph, 
making the message-passing types more diverse than simply retaining the one with the highest path strength. 
An alternative solution to derive a meta-multigraph is to set a threshold $\tau$, where message-passing types with path strengths above are retained.  
Here, we set the threshold $\tau^{(i,j)} $ as a value between the largest and the smallest path strengths in each multi-edge $\mathcal{R}^{(i,j)}$. We first briefly note the largest and smallest path strengths to clarify:
\begin{equation}
p^{(i,j)}_{max}=
\max\limits_{r \in \mathcal{R}^{(i,j)}}\{p^{(i,j)}_r\},
\label{eq:pmax}
\end{equation}
\begin{equation}
p^{(i,j)}_{min}= \min\limits_{r \in \mathcal{R}^{(i,j)}}\{p^{(i,j)}_r\},
\label{eq:pmin}
\end{equation}
where the path strength $p^{(i,j)}_r$ is calculated by Eq.~\ref{eq:softmax}. Then the threshold can be calculated as:
\begin{equation}
\tau^{(i,j)}= \lambda \cdot
p^{(i,j)}_{max}+ (1-\lambda) \cdot p^{(i,j)}_{min},
\label{eq:tau1}
\end{equation}
where $p^{(i,j)}_r$ is inherited from the search stage, $\lambda \in [0,1]$ is a hyper-parameter controlling the number of retrained paths in each multi-edge. $\lambda$ is a number closed to $1$ to ensure that all effective paths are retained and weak paths are dropped. 

Then, we formulate the information propagated from $\bm{H}^{(i)}$ to $\bm{H}^{(j)}$ in the derived meta-multigraph as an unweighted sum over candidate paths with path strengths above $\tau^{(i,j)}$:
\begin{equation}
\bm{H}^{(j)}=\sum_{i<j}
\sum_{r\in \hat{S}^{(i,j)}}{f}\left(\bm{\mathcal{A}}^{(i,j)}_r,\bm{H}^{(i)}\right),
\label{eval1}
\end{equation}
\begin{equation}
\hat{S}^{(i,j)}=\{r|p^{(i,j)}_r \ge \tau^{(i,j)},\forall r \in \mathcal{R}^{(i,j)}\},
\label{handcrafted}
\end{equation}
where $\hat{S}^{(i,j)}$ is the set of all retained paths between hyper-nodes $\bm{H}^{(i)}$ and $\bm{H}^{(j)}$. Then, the derived meta-multigraph can be used as the target-net for retraining from scratch.

\begin{figure*}[tb]
    \centering
    \includegraphics[width=0.82\linewidth]{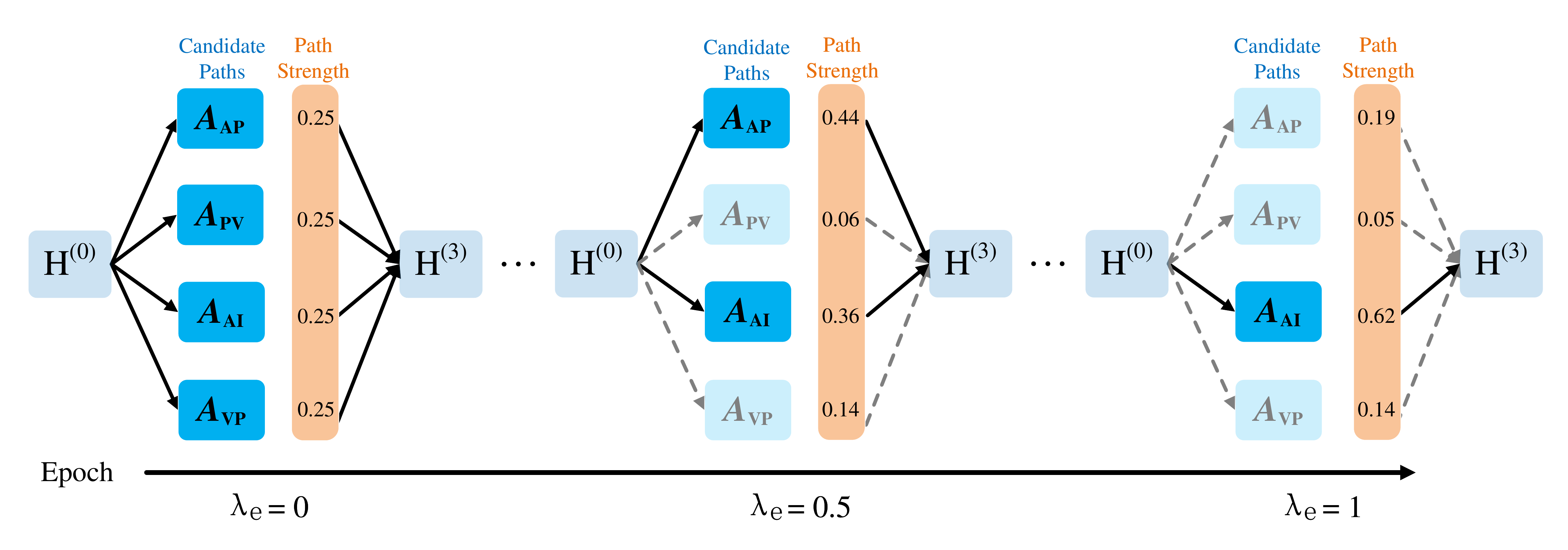} \\
    \vspace{-1ex}
    \caption{A diagram of the search algorithm of \model. We only show the search process between $\bm{H}^{(0)}$ and $\bm{H}^{(3)}$ for ease of illustration. The number of activated paths decreases with the increment of epochs. The collection of activated paths at the end of the search makes up the final target-net. 
    }
    \vspace{-0.5ex}
    \label{fig:progressive}
\end{figure*}
\subsection{Progressive Search Algorithm for Consistency and Stability}

In differentiable architecture search, the super-net trained in the search stage is a weighted summation among all candidate connections with a trainable distribution induced by softmax. It essentially optimizes a path combination. Yet path combination does not quite reach the actual goal: what really matters to the meta-structure search is to finish the path selection. 
After the search, the target-net is derived from the super-net by pruning weak paths. However, as noted by ~\cite{chang2019data,yao2020efficient} in the field of computer vision, connections may be highly correlated. Even if the path strength of some paths is small, the corresponding paths may be indispensable for performance. 
So the target-net derived from a high-performance super-net is not ensured to be good. 
The inconsistency will be detrimental to the performance. DiffMG randomly samples one candidate path on each multi-edge in each iteration, which reduces the inconsistency indirectly but results in high instability in the searched meta-graphs. Our PMMM addresses the instability issue of DiffMG but still contains a deriving step to get the final meta-structure, which leads to significant inconsistency between the super-net and the target-net. 
\begin{figure}[!t]
    \centering
    \subfloat[DARTS]{\includegraphics[width=0.45\linewidth]{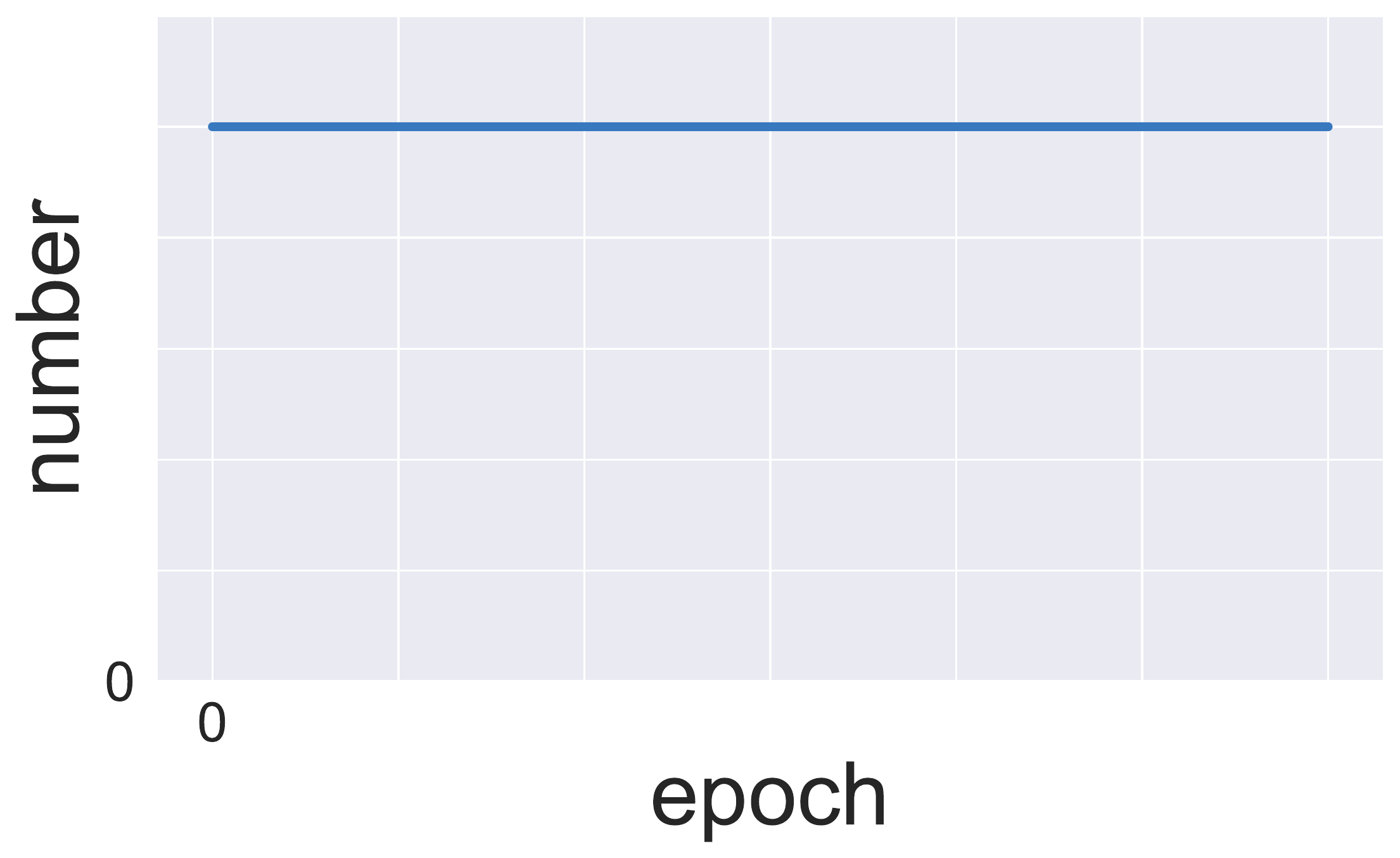}} 
    \hspace{2em}
    \subfloat[DiffMG]{\includegraphics[width=0.45\linewidth]{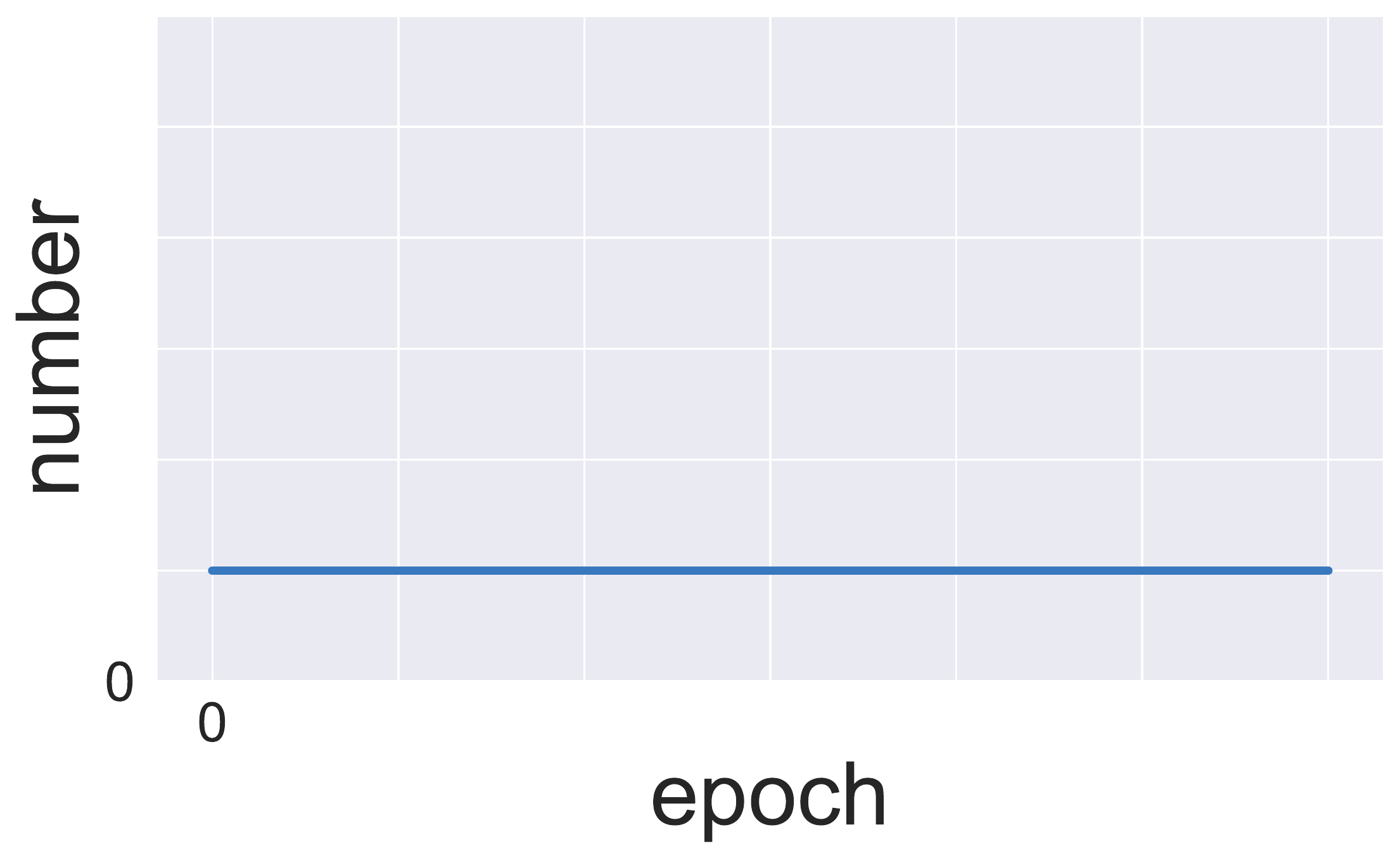}}
    \\
    \subfloat[$\lambda_e$]{\includegraphics[width=0.45\linewidth]{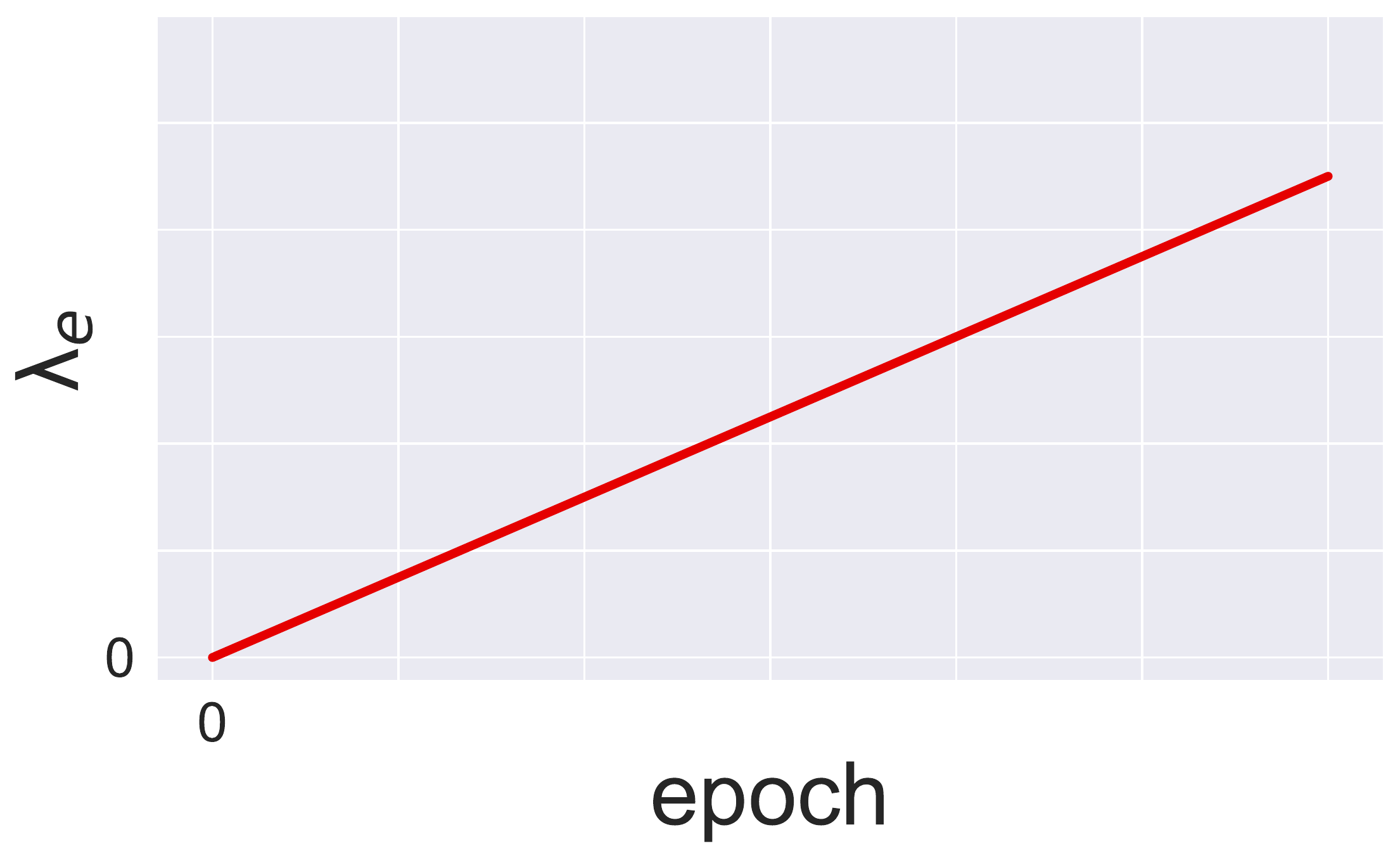}} 
    \hspace{2em}
    \subfloat[\model]{\includegraphics[width=0.45\linewidth]{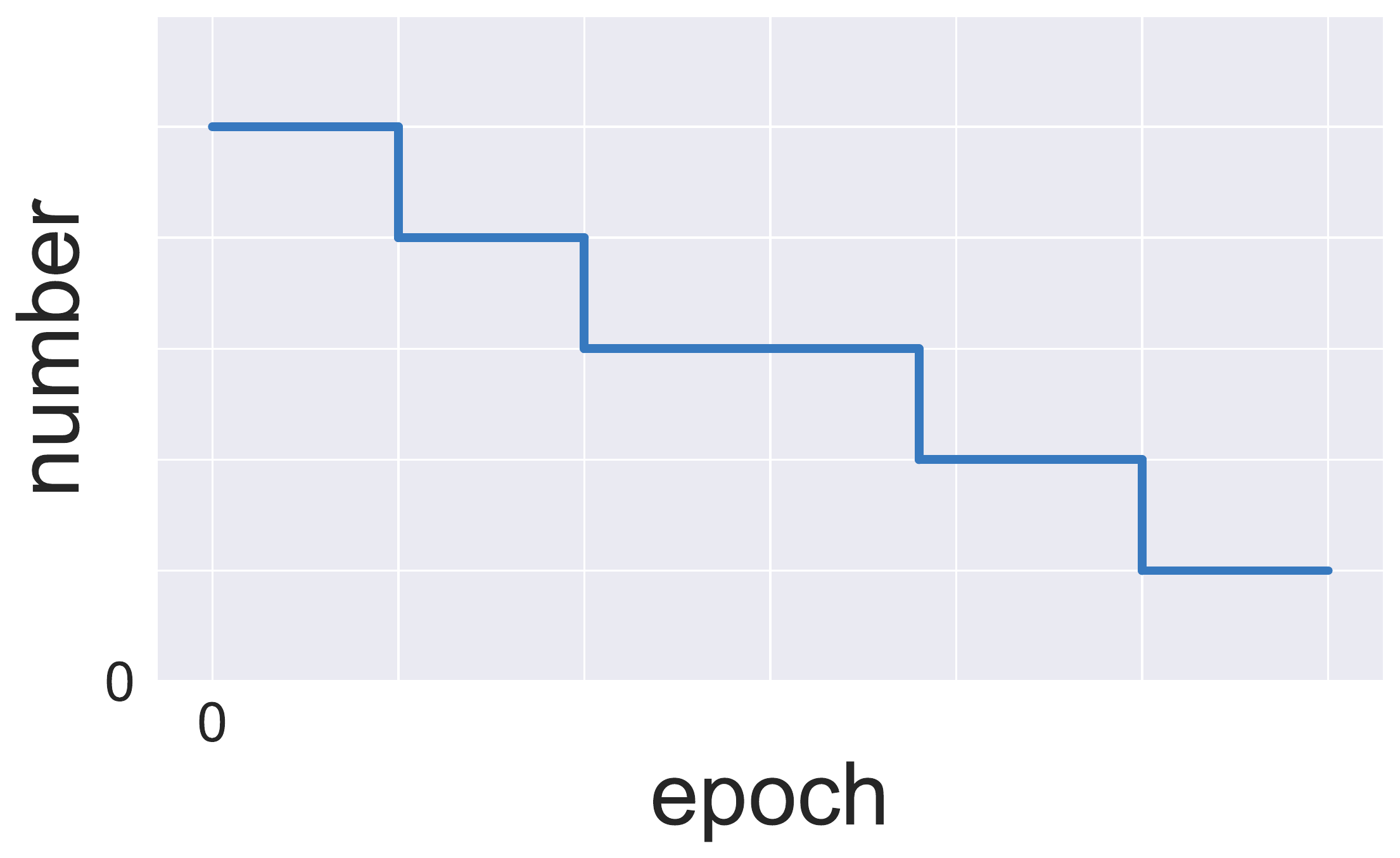}}
    \caption{Illustration of the change in the number of candidate paths over epochs in (a) DARTS, (b) DiffMG, and (d) \model, respectively. (c) shows the change of $\lambda_e$ over epochs in \model. The number of candidate paths in (d) decreases with $\lambda_e$ growing.
    }
    \vspace{-1ex}
    \label{fig:lambda}
\end{figure}

To address the inconsistency issue, we need to finish the path selection at the search stage so that the activated paths at the end of the search stage directly make up the target-net in the evaluation stage. To keep stability, the sampling probability of each path should not be very low.
Based on the above considerations, we propose a progressive 
search algorithm (\model) to gradually decrease the number of activated paths until the final activated paths make up the target-net.
Specifically, we only allow paths associated with the path strength larger than a threshold to be calculated to obtain $\partial \mathcal{L}_{\text{train}} / \partial \bm{\omega}$. With the threshold progressively increasing with the search epochs, the number of activated paths will be progressively reduced. Here, we set the threshold $\tau^{(i,j)} $ as a value between the largest path strengths $p^{(i,j)}_{max}$ and the smallest path strengths $p^{(i,j)}_{min}$ in multi-edge $\mathcal{R}^{(i,j)}$. As both $p^{(i,j)}_{max}$ and $p^{(i,j)}_{min}$ keep changing during the search stage, $\tau^{(i,j)} $ needs to be changed accordingly. So we set a hyper-parameter $\lambda_e\in [0,1]$ to control $\tau^{(i,j)} $. The threshold can be calculated as:
\begin{equation}
\tau^{(i,j)}_e= \lambda_e \cdot
p^{(i,j)}_{max}+ (1-\lambda_e) \cdot p^{(i,j)}_{min},
\label{eq:tau}
\end{equation}
where $\lambda_e$ 
is related to epoch $e$. As shown in Fig.~\ref{fig:lambda} (c) and (d), when $\lambda_e=0$, all paths are activated. When $\lambda_e=1$, only the strongest path is activated. At the search stage, $\lambda_e=0$ at the beginning and increases to $\lambda$, \ie, $1$. Correspondingly, $\tau^{(i,j)}_e$ increases with the search epochs. Fig.~\ref{fig:lambda} (a), (b) and (d) show the change in the number of candidate paths over epochs in different methods. DARTS always calculates all paths while DiffMG always samples one path. \model~could be viewed as a progressive transition from DARTS to DiffMG but more than a transition because \model~has completed the path selection in this process.

\begin{algorithm}[!t]
\caption{Search algorithm}
\label{alg}
\begin{algorithmic}[1]
\REQUIRE~~\\
Network weights $\bm{\omega}$; Architecture parameters $\bm{\alpha}$;  
Number of epochs $E$; 
Grow strategy for $\lambda_e, e \in \{1,2,\cdots,E\}$.
\ENSURE ~~\\
A meta-structure.
\STATE Initialize network weights $\bm{\omega}$ and architecture parameters $\bm{\alpha}$
\FOR{each epoch $e \in [1,E]$}
\STATE Get the threshold $\tau^{(i,j)}_e$ based on $\lambda_e$ and Eq.~\ref{eq:tau} 
\STATE Determine activated paths ${S}^{(i,j)}_e$ in each epoch based on Eq.~\ref{eq:active}. The collection of weights and architecture parameters of ${S}^{(i,j)}_e$ is denoted as $\hat{\bm{\omega}}$ and $\hat{\bm{\alpha}}$, respectively
\STATE Update weights $\hat{\bm{\omega}}$ by  $\nabla_{\hat{\bm{\omega}}}{\cal L}_{train}(\hat{\bm{\omega}}, \hat{\bm{\alpha}})$
\STATE Update parameters $\hat{\bm{\alpha}}$ by  $\nabla_{\hat{\bm{\alpha}}}{\cal L}_{val}(\hat{\bm{\omega}}, \hat{\bm{\alpha}})$
\ENDFOR 
\RETURN Generate the final meta-structure based on the activated paths.
\end{algorithmic}
	
\end{algorithm}

Based on $\tau^{(i,j)}_e$, we can get the set of all activated paths between hyper-node $\bm{H}^{(i)}$ and $\bm{H}^{(j)}$ during the search stage:
\begin{equation}
S^{(i,j)}_e=\{r \lvert p^{(i,j)}_r \ge \tau^{(i,j)}_e,\forall r \in \mathcal{R}^{(i,j)} \}.
\label{eq:active}
\end{equation}

Then we formulate the information propagated from $\bm{H}^{(i)}$ to $\bm{H}^{(j)}$ as a weighted sum over activated candidate message-passing steps: 
\begin{align}
\bm{H}^{(j)}=\sum_{i<j}
\sum_{r\in S^{(i,j)}_e}\bar{p}^{(i,j)}_r {Aggr}\left(\bm{\mathcal{A}}^{(i,j)}_r,\bm{H}^{(i)}\right),
\label{eq:forward_ours2}
\end{align}
\begin{align}
\bar{p}^{(i,j)}_r=\frac{\lvert \gR^{(i,j)}\rvert}{\lvert S^{(i,j)}_e\rvert} \cdot p^{(i,j)}_r,
\label{eq:hat_softmax}
\end{align}
where $p^{(i,j)}_r$ is calculated by Eq.~\ref{eq:softmax} and $\hat{p}^{(i,j)}_r$ is normalized to $p^{(i,j)}_r$. The overall algorithm is given in Algorithm 2.

\subsection{Search for Different Meta-structures}
\label{sec:meta}
 
Once the training of architecture parameters has been completed, we can then derive the compact meta-structure by pruning redundant paths. 
To investigate the relationship between the types of meta-structures and the performance of corresponding HGNNs, we search for four different meta-structures using our progressive search algorithm.

\subsubsection{Meta-path}
To integrate meta-path into the automatic framework, we expand the node type of Definition~\ref{def:path} into a hyper-node, generalizing a meta-path to a more flexible definition on the network schema. The meta-graph below is also extended.  

To search a meta-path, each multi-edge only retains edge types with the maximum path strength. Based on Eq.~\ref{eq:tau}, we set $\lambda=1$. So at the end of the search stage, $\lambda_e=1$ and  $\tau^{(i,j)}=p^{(i,j)}_{max}$.
In addition, all residual paths should be removed. So the input of $\bm{H}^{(j)}$ only comes from $\bm{H}^{(j-1)}$. We transform $\bm{H}^{(j)}$ in Eq.~\ref{eq:forward_ours2} into $\hat{\bm{H}}^{(j)}$ to formulate the propagated information:
\begin{align}
\hat{\bm{H}}^{(j)}=
\sum_{r\in S^{(j-1,j)}} \hat{p}^{(j-1,j)}_r {Aggr}\left(\bm{\mathcal{A}}^{(j-1,j)}_r,\bm{H}^{(j-1)}\right).
\label{eq:forward_path}
\end{align}

\subsubsection{Meta-graph}
Similar to generating a meta-path, we set $\lambda=1$ for a meta-graph. So $\tau^{(i,j)}=p^{(i,j)}_{max}$ at the end of the search stage. However, the residual paths are retained. The collection of paths between hyper-node $\bm{H}^{(i)}$ and $\bm{H}^{(j)}$ during the search stage can be directly determined by Eq.~\ref{eq:active}.

\subsubsection{Meta-multigraph}
A meta-multigraph allows propagating multiple message-passing types between two different hyper-nodes. To generate a meta-multigraph, each multi-edge may retain multiple edge types. We set $\lambda=0.9$ to ensure that all effective paths are retained and weak paths are dropped. 
At the end of the search stage, the threshold is a value close to the maximum path strength:
\begin{equation}
\tau^{(i,j)}= \lambda \cdot
p^{(i,j)}_{max}+ (1-\lambda) \cdot p^{(i,j)}_{min}.
\end{equation} 
The collection of paths between hyper-node $\bm{H}^{(i)}$ and $\bm{H}^{(j)}$ can be determined by Eq.~\ref{eq:active}.

\subsubsection{Complex-to-concise meta-multigraph}
Meta-multigraph is more flexible and complex than a meta-graph to represent more intricate semantic relations. However, the flexibility of meta-multigraphs may introduce 
redundant paths. For example, in Fig.~\ref{fig:multigraph} (d), a task-independent edge type $\bm{\mathcal{A}}_{PV}$ may connect $\bm{H}^{(0)}$ and $\bm{H}^{(2)}$ while only nodes of author type have labels. These 
redundant paths 
may have negative impacts on performance.



Intuitively, for a meta-structure, exploration is more important for the shallow parts while exploitation is more critical for the deep parts. So the message-passing types should be more gathered in deeper parts of the meta-multigraph to avoid task-independent edge types, where the tendency is from complex to concise.  
To generate a \meta~meta-multigraph, we need to adjust the threshold $\tau^{(i,j)}$ based on the depth of the meta-multigraph. So we transform ${\tau}^{(i,j)}$ in Eq.~\ref{eq:tau} into $\hat{\tau}^{(i,j)}$:
\begin{equation}
\hat{\tau}^{(i,j)} = \lambda^{j}_e \cdot
p^{(i,j)}_{max} + (1-\lambda^{j}_e) \cdot p^{(i,j)}_{min},
\label{eq:tau_p}
\end{equation}
\begin{equation}
\lambda^{j}_e =\beta^{N-j} \cdot \lambda_e,
\end{equation}
where $\beta \in [0,1]$ is a hyper-parameter controlling the attenuation ratio. We set $\beta=0.9$ and $\lambda=1$.  
So $\lambda_e=1$ at the end of the search stage and $\lambda^{j}_e$ declines as $j$ decreases. 
The collection of paths between hyper-nodes $\bm{H}^{(i)}$ and $\bm{H}^{(j)}$ can be determined by substituting $\tau^{(i,j)}$ in Eq.~\ref{eq:active} with $\hat{\tau}^{(i,j)}$:
\begin{equation}
S^{(i,j)}_e=\{r \lvert p^{(i,j)}_r \ge \hat{\tau}^{(i,j)}_e,\forall r \in \mathcal{R}^{(i,j)} \}.
\label{eq:active_t}
\end{equation}

For the above four meta-structures, we can directly obtain the corresponding target-nets by letting $\lambda^{(i,j)}_e=\lambda$ in  Eq.~\ref{eq:active}. Then, the information propagated from $\bm{H}^{(i)}$ to $\bm{H}^{(j)}$ in the evaluation stage can be formulated as
an unweighted-version of Eq.~\ref{eq:forward_ours2}:
\begin{align}
\bm{H}^{(j)}=\sum_{i<j}
\sum_{r\in S^{(i,j)}_e} {Aggr}\left(\bm{\mathcal{A}}^{(i,j)}_r,\bm{H}^{(i)}\right).
\label{eq:eval2}
\end{align}


\subsection{Differences to Prior Works}
\begin{table}[t]
	\centering
	\caption{Comparison of PMMM with related NAS algorithms. \textit{Coupling} denotes whether all architecture parameters are coupling optimized. \textit{Probability} denotes the updating probability of each path in one iteration.
		}
	\label{tab:search_and_derive}
	\begin{threeparttable}[b]
	\small
	\resizebox{0.45\textwidth}{!}{
	\begin{tabular}{l  c c c}
		\toprule
		\multirow{2}{*}{Method}       & 
		\multicolumn{2}{c}{Search}  
		 &  \multirow{2}{*}{Derivation}    \\ 
		&Coupling&Probability&  \\
		\midrule
		DARTS~\cite{DBLP:conf/iclr/LiuSY19}     
		& \ding{51}   &      high       &    single path        \\
		P-DARTS~\cite{chen2019progressive}     
		& \ding{51}   &      high       &    single path        \\
		PC-DARTS~\cite{xu2019pc}
		&  \ding{51}  &      high       &    single path        \\
		ProxylessNAS~\cite{xu2019pc}
		&  \ding{55}  &      low       &    single path        \\
		SPOS~\cite{guo2020single}
		&  \ding{55}   &     low        &   single path         \\
		DiffMG~\cite{yao2019differentiable}
		&   \ding{55}    &   low          &   single path         \\
		\midrule
		PMMM       & \ding{55}    &   middle          &   multiple paths         \\ 
		\model       & \ding{55}    &   high$\rightarrow$low          &   multiple paths         \\ \bottomrule
	\end{tabular}}
	\end{threeparttable}
\end{table}
Table~\ref{tab:search_and_derive} details the differences between our approach and related differentiable NAS algorithms, including DARTS \cite{DBLP:conf/iclr/LiuSY19}, P-DARTS~\cite{chen2019progressive},  PC-DARTS~\cite{xu2019pc}, ProxylessNAS~\cite{cai2018proxylessnas}, SPOS \cite{guo2020single}, and DiffMG~\cite{yao2019differentiable}. 
The first five algorithms search convolution or pooling operations in CNNs instead of meta-structures in GNNs. We ignore these differences and focus on the algorithms.
In contrast to DARTS, P-DARTS, and PC-DARTS, our search algorithms do not jointly optimize all the architecture parameters, which reduces the inconsistency between the search phase and the evaluation phase. Compared to ProxylessNAS, SPOS, and DiffMG, our search algorithms update each message-passing step with a higher probability,   
avoiding unfairness caused by insufficient training. Regarding the derivation strategy, our approaches are distinct from all the above methods. \par


The most related NAS method to \model~is P-DARTS \cite{chen2019progressive}, which is employed in the computer vision field. It reduces the search space by splitting the search process into three stages and  explicitly removing a part of less important connections at the end of the first two stages. The main drawback is that candidate connections discarded early may be powerful at the end of the search, which means useful connections may be dropped too early.



\section{Experiments}
\label{sec:experiments}
\begin{table}[!t]
\caption{Statistics of datasets used in this paper.} 
\resizebox{\linewidth}{!}{
\begin{tabular}{ccccc}
\hline
Dataset  & \#Nodes   & \begin{tabular}[c]{@{}c@{}}\#Node\\ types\end{tabular} & \#Edges    & \#Classes \\ \hline
DBLP     & 26,128    & 4 & 239,566    & 4   \\
ACM      & 10,942    & 4 & 547,872    & 3   \\
IMDB     & 21,420    & 4 & 86,642     & 5   \\
Ogbn-mag & 1,939,743 & 4 & 21,111,007 & 349 \\ \hline
\end{tabular}
}
\label{tab:s_nc}
\vspace{-0.5ex}
\end{table}

\begin{table}[!t]
	\small
	\centering
	\caption{Statistics of HINs for recommendation.}
    \label{tab:s_re}
	\begin{threeparttable}[b]
	\small
	\resizebox{0.48 \textwidth}{!}{
	\begin{tabular}{lcccc 
	}
		\toprule
		Dataset                   &  Relations (A-B)                & \#~A           & \#~B             & \#~A-B           \\ \midrule
		\multirow{5}{*}{Yelp}     &  User-Business (U-B)   & 16239 & 14284   & 198397  \\
		~                         &  User-User (U-U)                & 16239          & 16239            & 158590           \\
		~                         &  User-Compliment (U-Co)         & 16239          & 11               & 76875            \\
		~                         &  Business-City (B-C)            & 14284          & 47               & 14267            \\
		~                         &  Business-Category (B-Ca)       & 14284          & 511              & 40009            \\ \midrule
		\multirow{6}{*}{\shortstack{Douban\\movie}} & User-Movie (U-M)  & 13367 & 12677 & 1068278 \\ 
		~   & User-Group (U-G) & 13367 & 2753  & 570047 \\
		~   & User-User  (U-U) & 13367 & 13367 & 4085  \\
		~   & Movie-Actor (M-A) & 12677 & 6311 & 33587 \\
		~   & Movie-Director (M-D) & 12677 & 2449 & 11276 \\
		~   & Movie-Type  (M-T) & 12677 & 38 & 27668 \\ \midrule
		\multirow{4}{*}{Amazon} & User-Item (U-I) & 6170 & 2753 & 195791 \\
		~ & Item-View (I-V) & 2753 & 3857 & 5694 \\
		~ & Item-Category (I-C) & 2753 & 22 & 5508 \\
		~ & Item-Brand (I-B) & 2753 & 334 & 2753 \\ \bottomrule
	\end{tabular}
	}
	\end{threeparttable}
\vspace{-1ex}
\end{table}
In this section, we first compare our \mone~and \model\ 
with state-of-the-art methods on seven datasets of node classification and recommendation tasks to 
evaluate their performance. Then, we show the efficiency and consistency of our methods, and analyze the stability of \mone~and \model. In addition, we visualize our searched meta-structures and visualize the results on DBLP. 
In the end, we conduct ablation studies on critical modules and hyper-parameters of our methods.
\begin{table*}[!t]
\caption{Macro-F1 ($\%$) and Micro-F1 ($\%$) on the node classification task (mean ± std).  
\textit{avg.} is obtained on 5 independently searched meta-structures to explore the stability. The best and second-best results are shown in bold and underlined, respectively.
}
\label{tab:nc}
\centering
\begin{threeparttable}[b]
\small
\resizebox{0.9\textwidth}{!}{
\begin{tabular}{lcccccc}
\toprule
   \multirow{2}{*}{Methods}                &  \multicolumn{2}{c}{IMDB}                      &  \multicolumn{2}{c}{DBLP}                      &  \multicolumn{2}{c}{ACM}                                         \\ 
                   &  Macro-F1              &  Micro-F1              &  Macro-F1              &  Micro-F1              &  Macro-F1              &  Micro-F1                           \\ \midrule

GCN	& $	57.88_{\pm1.18	} $ & $	64.82_{\pm0.64	} $ & $	90.84_{\pm0.32	} $ & $	91.47_{\pm0.34	} $ & $	92.17_{\pm0.24	} $ & $	92.12_{\pm0.23	} $ \\
GAT	& $ 58.94_{\pm1.35	} $ & $	64.86_{\pm0.43	} $ & $	93.83_{\pm0.27	} $ & $	93.39_{\pm0.30	} $ & $	92.26_{\pm0.94	} $ & $	92.19_{\pm0.93	} $ \\
RGCN	& $	58.85_{\pm0.26	} $ & $	62.05_{\pm0.15	} $ &$	91.52_{\pm0.50	} $ & $	92.07_{\pm0.50	} $ &  $	91.55_{\pm0.74	} $ & $	91.41_{\pm0.75	} $ \\
HAN	& $	57.74_{\pm0.96	} $ & $	64.63_{\pm0.58	} $ & $	91.67_{\pm0.49	} $ & $	92.05_{\pm0.62	} $ & $	90.89_{\pm0.43	} $ & $	90.79_{\pm0.43	} $  \\
GTN	& $	60.47_{\pm0.98	} $ & $	65.14_{\pm0.45	} $ & $	93.52_{\pm0.55	} $ & $	93.97_{\pm0.54	} $ & $	91.31_{\pm0.70	} $ & $	91.20_{\pm0.71	} $ \\
HetGNN	& $	48.25_{\pm0.67	} $ & $	51.16_{\pm0.65	} $ & $	91.76_{\pm0.43	} $ & $	92.33_{\pm0.41	} $ & $	85.91_{\pm0.25	} $ & $	86.05_{\pm0.25	} $ \\
MAGNN	& $	56.49_{\pm3.20	} $ & $	64.67_{\pm1.67	} $ & $	93.28_{\pm0.51	} $ & $	93.76_{\pm0.45	} $ & $	90.88_{\pm0.64	} $ & $	90.77_{\pm0.65	} $ \\
HGT	& $	63.00_{\pm1.19	} $ & $	67.20_{\pm0.57	} $ & $	93.01_{\pm0.23	} $ & $	93.49_{\pm0.25	} $ & $	91.12_{\pm0.76	} $ & $	91.00_{\pm0.76	} $ \\ 
Simple-HGN & $ 63.53_{\pm1.36}  $ & $ 67.36_{\pm0.57}  $ & $ 94.01_{\pm0.24}  $ & $ 94.46_{\pm0.22}  $ & $ 93.42_{\pm0.44}  $ & $ 93.35_{\pm0.45}  $  \\
DiffMG & $ 49.35_{\pm1.89}  $ & $ 59.75_{\pm1.23}  $ & $ 93.68_{\pm0.37}  $ & $ 94.20_{\pm0.36}  $ & $ 88.03_{\pm3.25}  $ & $ 88.07_{\pm3.04}  $   \\
\mone~(ours) & $ \underline{64.49_{\pm0.39}} $ & $ \underline{68.26_{\pm0.22}}  $ & $ \underline{94.79_{\pm0.14}}  $ & $ \underline{95.15_{\pm0.15}}  $ & $\underline{93.79_{\pm0.16} } $ & $ \underline{93.71_{\pm0.17}}  $    \\
\model~(ours) & $\bm {64.92_{\pm0.23}} $ & $ \bm {68.74_{\pm0.44}}  $ & $ \bm {94.86_{\pm0.13}}  $ & $ \bm {95.26_{\pm0.12}}  $ & $\bm { 94.02_{\pm0.21}}  $ & $ \bm {93.98_{\pm0.22}}  $    \\
\midrule
\midrule
DiffMG (avg.) & $ 41.89_{\pm4.73}  $ & $ 55.98_{\pm2.54}  $ & $ 89.17_{\pm4.84}  $ & $ 89.82_{\pm4.69}  $ & $ 86.12_{\pm9.18}  $ & $ 86.46_{\pm8.37}  $  \\
\mone~ (avg.) & $  63.89_{\pm1.49}  $ & $ 67.58_{\pm1.27}  $ & $ \bm {94.67_{\pm0.24}}  $ & $ \bm {95.14_{\pm0.22}}  $ & $ 93.19_{\pm0.63}  $ & $ 93.12_{\pm0.64}  $  \\
\model~ (avg.) & $ \bm{64.11_{\pm0.49}}  $ & $ \bm {68.19_{\pm0.76}}  $ & $ 94.33_{\pm0.43}  $ & $ 94.71_{\pm0.43}  $ & $ \bm {93.62_{\pm0.44}}  $ & $ \bm {93.52_{\pm0.46}}  $  \\
\bottomrule
\end{tabular}
}
\end{threeparttable}
\end{table*}

\subsection{Experimental Setup}
\label{sec:setup}
\subsubsection{Datasets}
We evaluate our methods on two popular tasks~\cite{yang2020heterogeneous}: node classification and recommendation. The node classification task aims to predict the correct labels for nodes based on network structure and node features. We use three widely-used real-world datasets: 
IMDB\footnote{\url{https://www.kaggle.com/karrrimba/movie-metadatacsv}}, DBLP\footnote{\url{http://web.cs.ucla.edu/~yzsun/data/}}, ACM~\cite{wang2019heterogeneous}, and a large-scale real-world dataset from OGB challenge~\cite{hu2021ogb}: ogbn-mag. 
For the recommendation task, we aim to predict links between source nodes (\eg, users) and target nodes (\eg, items). We adopt three widely-used heterogeneous recommendation datasets\footnote{~In\url{https://github.com/librahu/HIN-Datasets-for-Recommendation-and-Network-Embedding}}: Amazon, Yelp, and Douban Movie (abbreviated as Douban).

As Simple-HGN and DiffMG are the most important baselines of our methods, we use the dataset splits in Simple-HGN for node classification on medium-scale datasets and the dataset splits in DiffMG for recommendation to avoid inappropriate preference settings. 
The details of all datasets are listed in Tables~\ref{tab:s_nc} and~\ref{tab:s_re}.


\subsubsection{Baselines}
We compare \mone~and \model~with fourteen methods, including: 
1) three homogeneous GNNs, \ie, GCN~\cite{kipf2016semi}, GAT~\cite{velivckovic2017graph}, and GAMLP~\cite{GAMLP2021} for large-scale ogbn-mag dataset;  
2) nine heterogeneous GNNs, \ie, RGCN~\cite{DBLP:conf/esws/SchlichtkrullKB18},  HetGNN~\cite{zhang2019heterogeneous},  HAN~\cite{wang2019heterogeneous}, GTN~\cite{NIPS2019_9367}, MAGNN~\cite{fu2020magnn}, HGT~\cite{hu2020heterogeneous}, Simple-HGN~\cite{lv2021we}, NARS~\cite{NARS2020} and SAGN~\cite{SAGN2021} for large-scale ogbn-mag dataset;  
3) two AutoML methods, \ie, GEMS~\cite{han2020genetic} for recommendation, and DiffMG~\cite{DBLP:conf/kdd/DingYZZ21}.

\subsubsection{Parameter Settings}
Following DiffMG, we run the search algorithm with different random search seeds three times to derive the meta-structure from the run that achieves the best validation performance. 
For a fair comparison, the parameter settings of \mone~and \model~are the same with DiffMG. 
We use Adam to train $\alpha$ with a learning rate of $3e-4$ and $\omega$ with a learning rate of 0.0005, setting the weight decay to 0.001. 
To coordinate with baselines, we set the steps (depth) $N = 4$, which is the same as the length of meta-paths learned by GTN and meta-graph searched by DiffMG. For PMMM, we set $p = 2$ for most datasets, \ie, only $1/2$ paths are randomly sampled on each edge, except for DBLP and ACM with small $K$, we set $p = 1$. For \model, we set $\lambda = 1$ and $\beta = 0.9$. The baseline scores come from Simple-HGN~\cite{lv2021we} for node classification and DiffMG~\cite{DBLP:conf/kdd/DingYZZ21} for the recommendation task. 

\subsubsection{Evaluation Metrics}
For evaluation, we use the Macro-F1 score as well as the Micro-F1 score as metrics for the medium-scale datasets, validation accuracy and test accuracy for the large-scale datasets on the node classification task, and AUC (area under the ROC curve) for the recommendation task. Due to the randomness of initialization, following DiffMG, we run the search algorithms of DiffMG, \mone~and \model~for $5$ times with different random search seeds to derive the meta-structures from the run that achieves the best validation performance.  
We evaluate all methods with different random training seeds for $5$ runs on node classification as in Simple-HGN and $10$ runs on recommendation as in DiffMG.

\begin{table}[!t]
\caption{Experiment results on the large-scale dataset ogbn-mag compared with other methods on the OGB leaderboard. 
}
\resizebox{\linewidth}{!}{
\begin{tabular}{lcc}
\toprule
Methods               & Validation accuracy & Test accuracy       \\ 
\hline
RGCN                  &$ 48.35_{\pm0.36          }$&$ 47.37_{\pm0.48          }$\\
HGT                   &$ 49.89_{\pm0.47          }$&$ 49.27_{\pm0.61          }$\\
NARS                  &$ 51.85_{\pm0.08          }$&$ 50.88_{\pm0.12          }$\\
SAGN                  &$ 52.25_{\pm0.30          }$&$ 51.17_{\pm0.32          }$\\
GAMLP                 &$ 53.23_{\pm0.23          }$&$ 51.63_{\pm0.22          }$\\ 
PMMM (ours)                 &$ \underline{54.25_{\pm0.17          }}$&$ \underline{52.87_{\pm0.21          }}$\\
\model~(ours)                &$ \bm{54.78_{\pm0.15          }}$&$ \bm{53.13_{\pm0.20          }}$\\ 

\bottomrule
\end{tabular}}
\label{tab:result_on_large_dataset}
\end{table}
\begin{table}[!t]
	\centering
	\caption{AUC ($\%$) on the recommendation task (mean ± std). }
	\label{tab:lp}
	\begin{threeparttable}[b]
	\small
	\resizebox{0.48\textwidth}{!}{
	\begin{tabular}{lccc 
	}
		\toprule
	Methods  & Amazon &  Yelp &  Douban \\
		\midrule
		GCN     & $ 66.64_{\pm1.00 } $ & $ 58.98_{\pm0.52  } $ & $ 77.95_{\pm0.05 } $ \\
        GAT     & $ 55.70_{\pm1.13   } $ & $ 56.55_{\pm0.05    } $ & $ 77.58_{\pm0.33 } $ \\
        HAN     & $ 67.35_{\pm0.11   } $ & $ 64.28_{\pm0.20    } $ & $ 82.65_{\pm0.08 } $ \\
        GTN     & $ 71.82_{\pm0.18   } $ & $ 66.27_{\pm0.31    } $ & $ 83.26_{\pm0.10 } $ \\
        MAGNN   & $ 68.26_{\pm0.09   } $ & $ 64.73_{\pm0.24    } $ & $ 82.44_{\pm0.17 } $ \\
        HGT     & $ 74.75_{\pm0.08 } $ & $ 68.07_{\pm0.35    } $ & $ 83.38_{\pm0.06 } $ \\
        GEMS   & $  70.66_{\pm0.14   } $ & $ 65.12_{\pm0.27    } $ & $ 83.00_{\pm0.05 } $ \\
        Simple-HGN  & $ 75.30_{\pm0.51 } $ & $ 68.92_{\pm0.32    } $ & $ 83.41_{\pm0.22 } $ \\
        DiffMG  & $ 75.28_{\pm0.08 } $ & $ 68.77_{\pm0.13    } $ & $ 83.78_{\pm0.09 } $ \\
        \mone~(ours)  & $  \underline{75.48_{\pm0.04} } $ & $ \underline{69.27_{\pm0.13}}  $ & $  \underline{83.88_{\pm0.06}}  $ \\
        \model~(ours)  & $ \bm{75.55_{\pm0.07  }} $ & $ \bm {69.54_{\pm0.08} } $ & $  \bm {83.99_{\pm0.03}}  $ \\
        \midrule
        \midrule
        DiffMG (avg.)& $ 74.08_{\pm0.96 } $ & $ 66.87_{\pm1.92    } $ & $ 83.06_{\pm0.85 } $ \\
        \mone~ (avg.) & $ 75.09_{\pm0.29 } $ & $  68.78_{\pm0.53    } $ & $ \bm {83.72_{\pm0.13 }} $ \\
        \model~ (avg.) & $ \bm {75.14_{\pm0.11 }} $ & $ \bm {68.84_{\pm0.21}} $ & $ 83.68_{\pm0.11 } $ \\
		
		\bottomrule
	\end{tabular}}
	\end{threeparttable}
\end{table}

\subsection{Comparison on Node Classification}
\subsubsection{Results on Medium-scale Datasets}
\label{sec:nc}
Table~\ref{tab:nc} compares the results of the proposed model, \mone~and \model~to the baseline scores from the HGB paper on node classification. 
Based on the results, we have several observations. 
First, Heterogeneous GNNs relying on manually designed meta-paths like HAN and MAGNN do not achieve desirable performance and perform worse than homogeneous GNNs like GAT, suggesting that hand-crafted rules may have adverse implications. 
Second, \mone~and \model~consistently outperform all the advanced baselines, demonstrating that HGNNs can significantly benefit from meta-structures if they are automatic and expressive. 
To explore the stability, we show average results (\textit{avg.}) of DiffMG, \mone~and \model~on $5$ independently searched models from different search seeds. The results show that both \mone~and \model~greatly surpass DiffMG in terms of stability.  

\subsubsection{Experiments on Large-scale Dataset}
For the ogbn-mag dataset, we use the official data partition, where papers published before 2018, in 2018, and since 2019 are nodes for training, validation, and testing, respectively. The experiments are conducted on a Tesla V100-PCIE 16GB GPU and the batch size is $10000$. 
The results are compared to the OGB leaderboard, and all scores are the average of $10$ separate training.
Results on Table~\ref{tab:result_on_large_dataset} show that \mone~and \model~achieve superior performance over other methods. 

\subsection{Comparison on Recommendation}
Table~\ref{tab:lp} reports the results of our methods and the baselines on the recommendation task. 
Both our \mone~and \model~surpass the state-of-the-art models on all three datasets, demonstrating that a meta-multigraph surpasses a meta-path or meta-graph in capturing the semantic information on HINs. The advantages of \mone~and \model~over DiffMG are more significant on node classification tasks than on recommendation tasks, showing the better adaptability of our methods. To investigate the stability of our methods, we show average results (\textit{avg.}) of DiffMG, \mone~and \model~on $5$ independently searched models from different search seeds. The results show that both \mone~and \model~greatly outperform DiffMG. \mone~and \model~also surpass HGT~\cite{hu2020heterogeneous}, a far more complex architecture, indicating the superiority of NAS-based methods in HINs. The results indicate the superiority of NAS-based methods in HINs.


\begin{figure}[!t]
    \centering
    \includegraphics[width=1\linewidth]{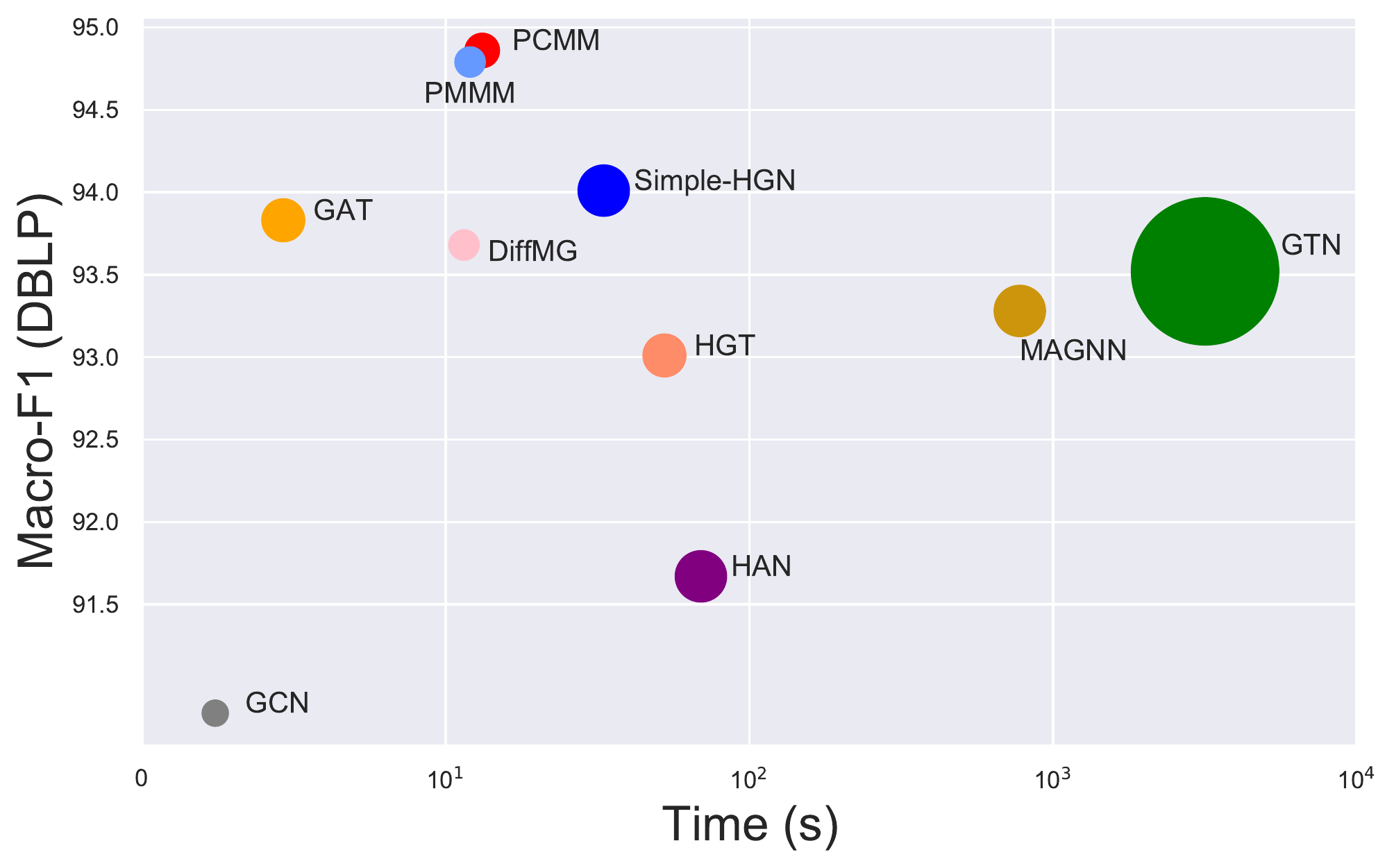}
    \caption{Time and memory comparison for HGNNs on
DBLP. The area of the circles represents the (relative) memory consumption of the corresponding models. Note that the abscissa increases exponentially.}
    \vspace{-1ex}
    \label{fig:efficiency}
\end{figure}
\subsection{Efficiency}
We evaluate the training cost and memory consumption of all available HGNNs for node classification on the DBLP dataset. The results are illustrated in Fig.~\ref{fig:efficiency}. The area of the circles represents the (relative) memory consumption of the corresponding models. We see that Simple-HGN shows a slightly better Macro-F1 score than GAT but much longer training cost. DiffMG has smaller memory consumption than Simple-HGN but worse performance. \mone~and \model~have the best performance and acceptable training cost and memory consumption. The automatic methods, DiffMG, \mone~and \model, have fairly small memory consumption (only larger than GCN),  
indicating that they can precisely propagate messages in a proper position to avoid extra computational overhead.

\begin{figure*}[!t]
\centering
\subfloat[IMDB]{\includegraphics[width=0.30\linewidth]{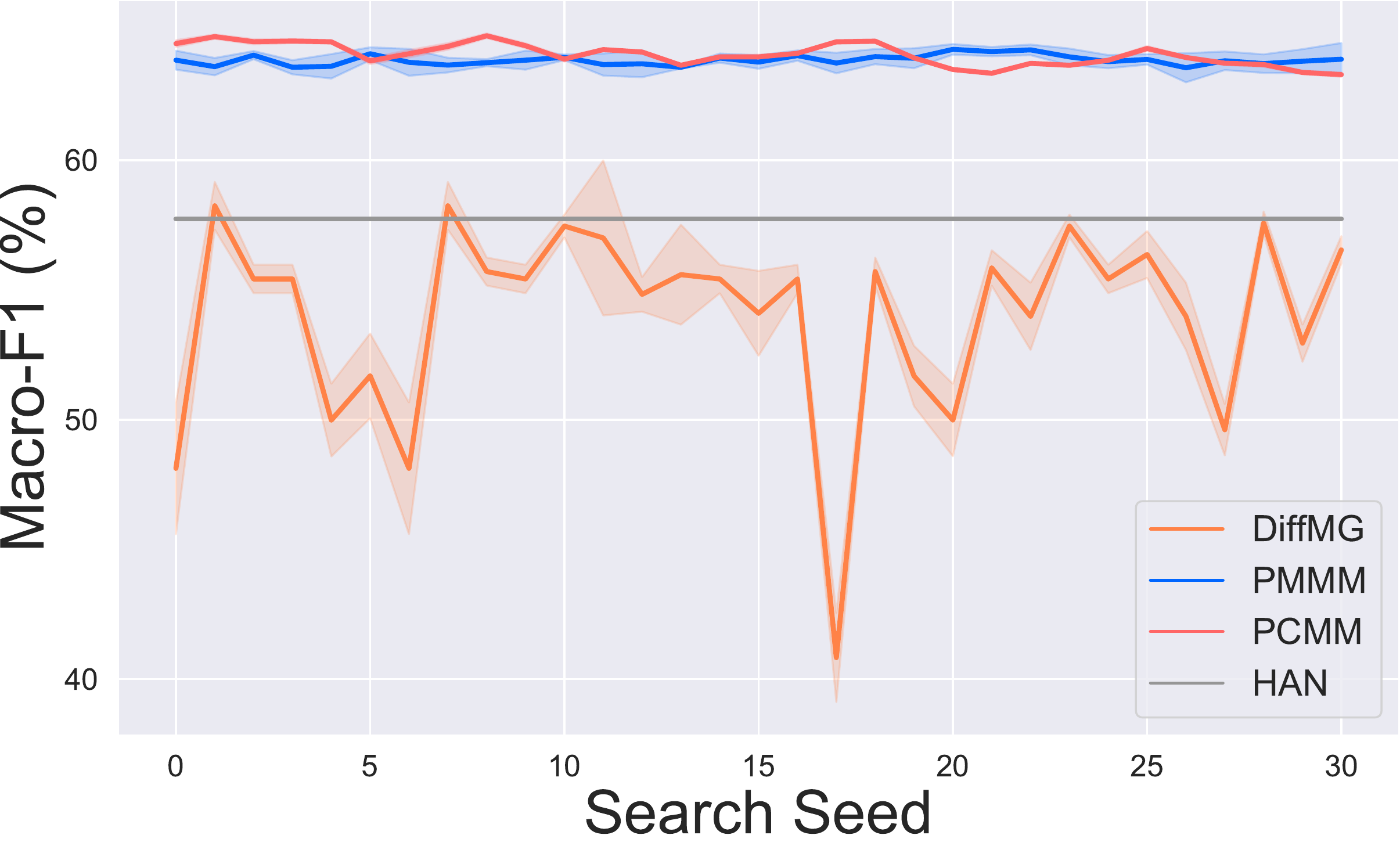}}
\hspace{1.0em}
\subfloat[DBLP]{\includegraphics[width=0.30\linewidth]{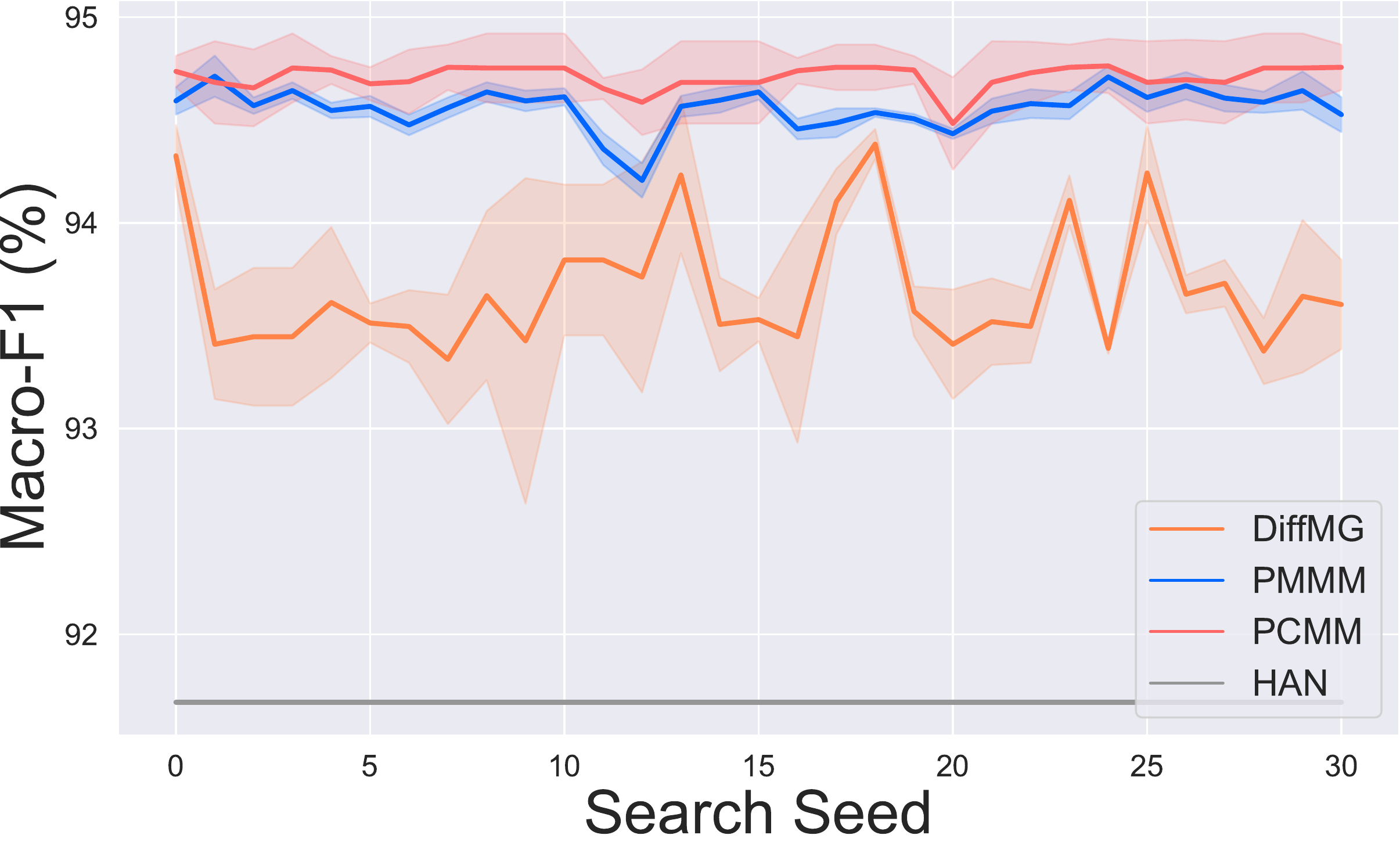}}
\hspace{1.0em}
\subfloat[ACM]{\includegraphics[width=0.30\linewidth]{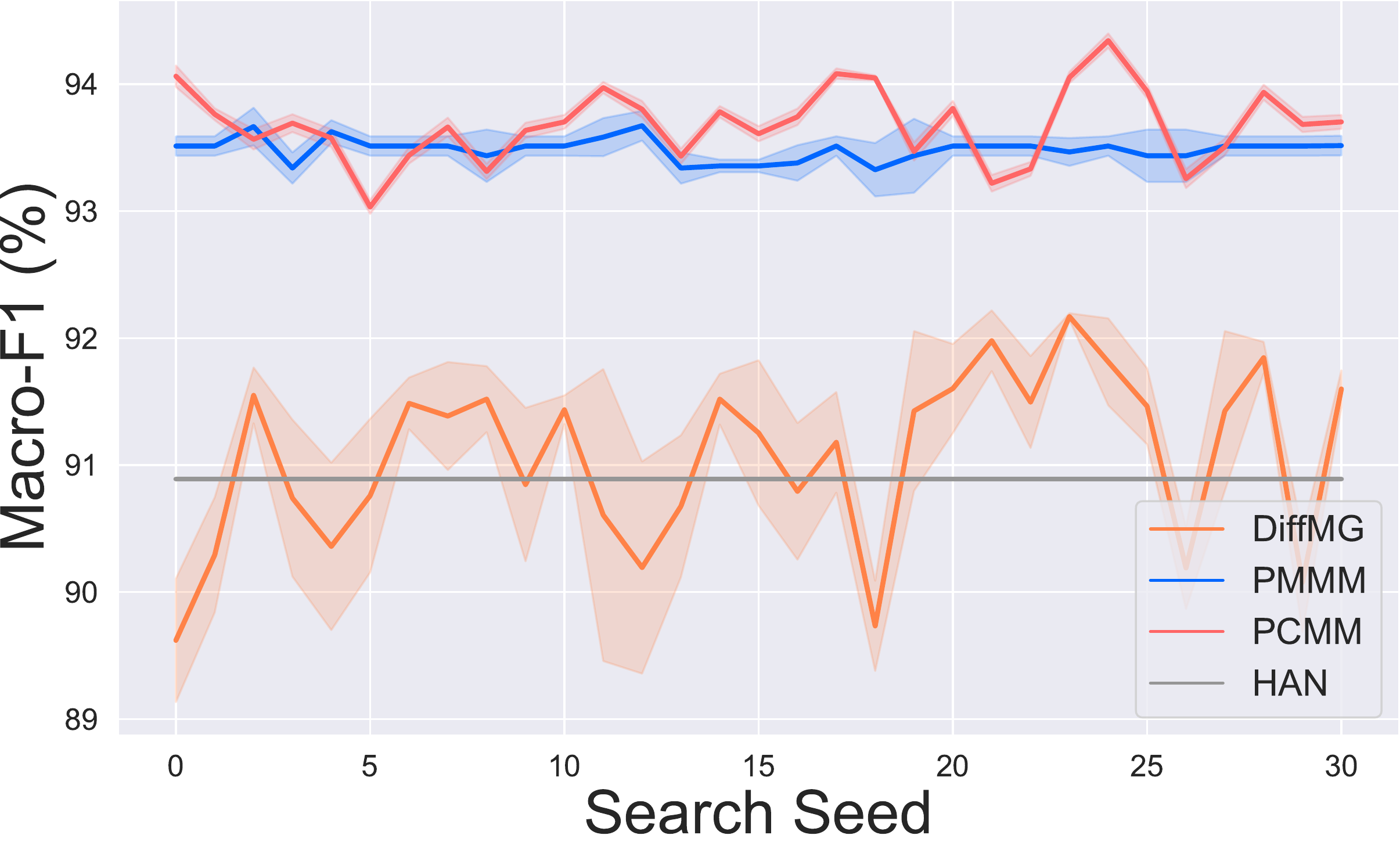}}

\subfloat[Amazon]{\includegraphics[width=0.30\linewidth]{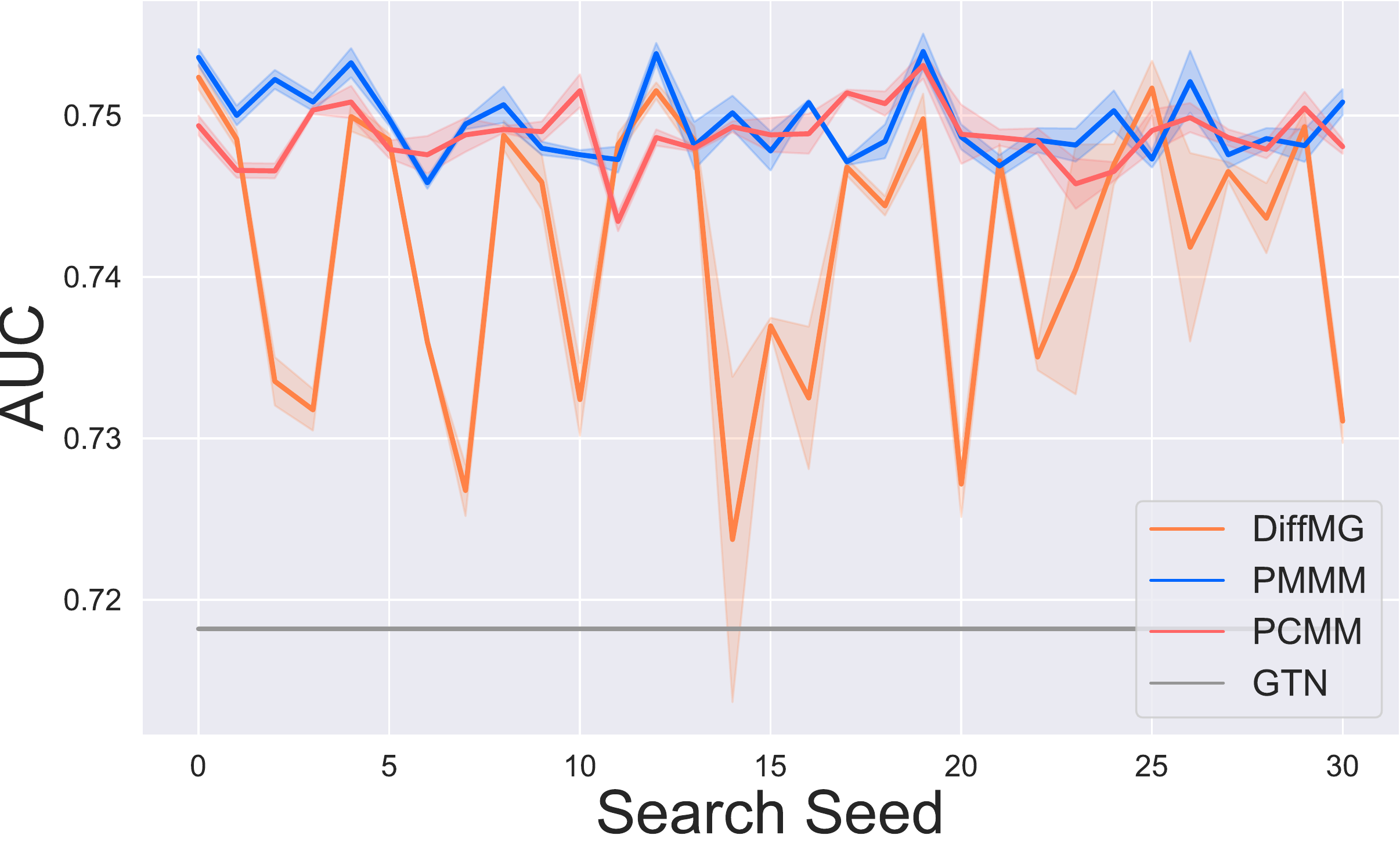}}
\hspace{1.0em}
\subfloat[Yelp]{\includegraphics[width=0.30\linewidth]{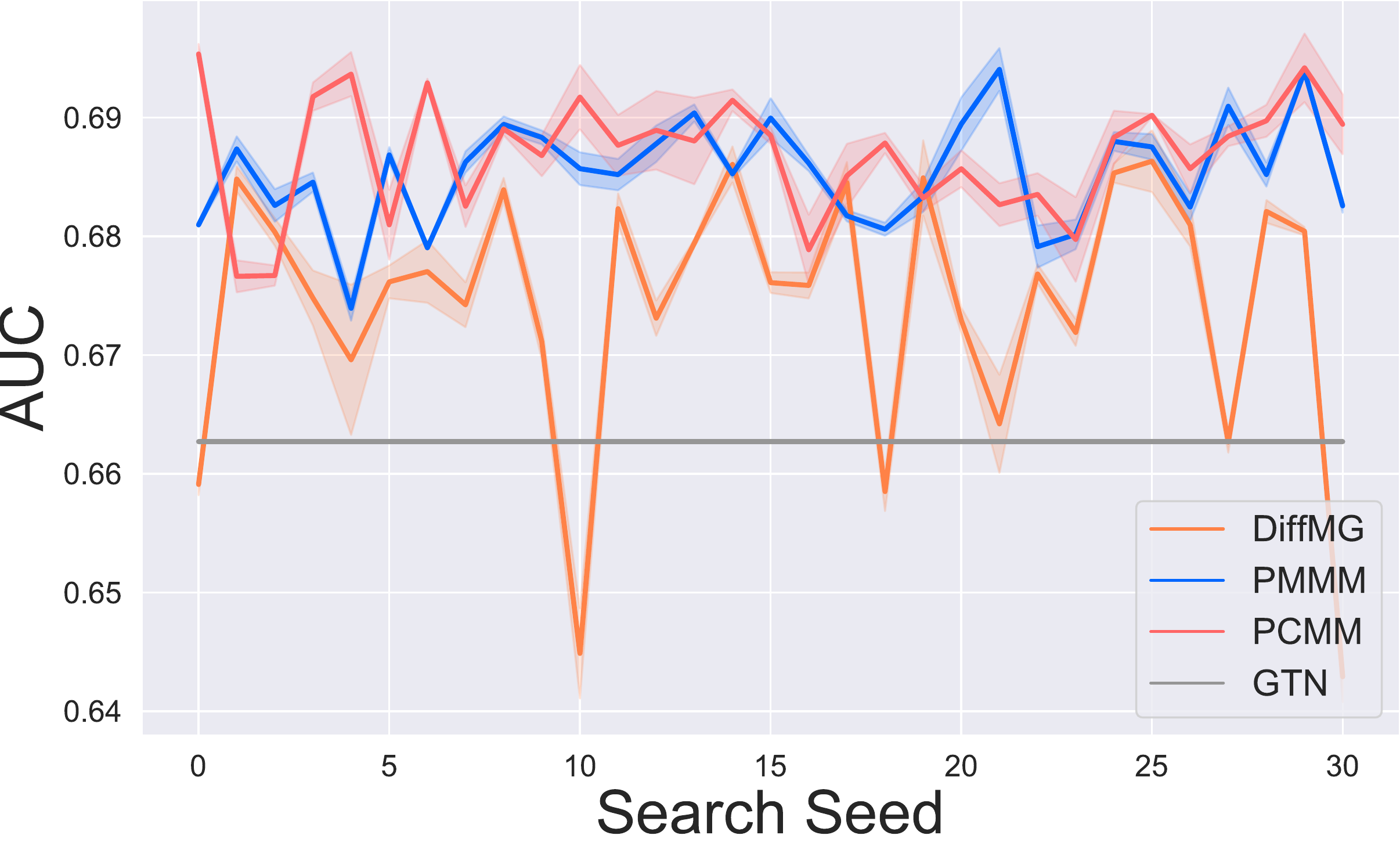}}
\hspace{1.0em}
\subfloat[Douban]{\includegraphics[width=0.30\linewidth]{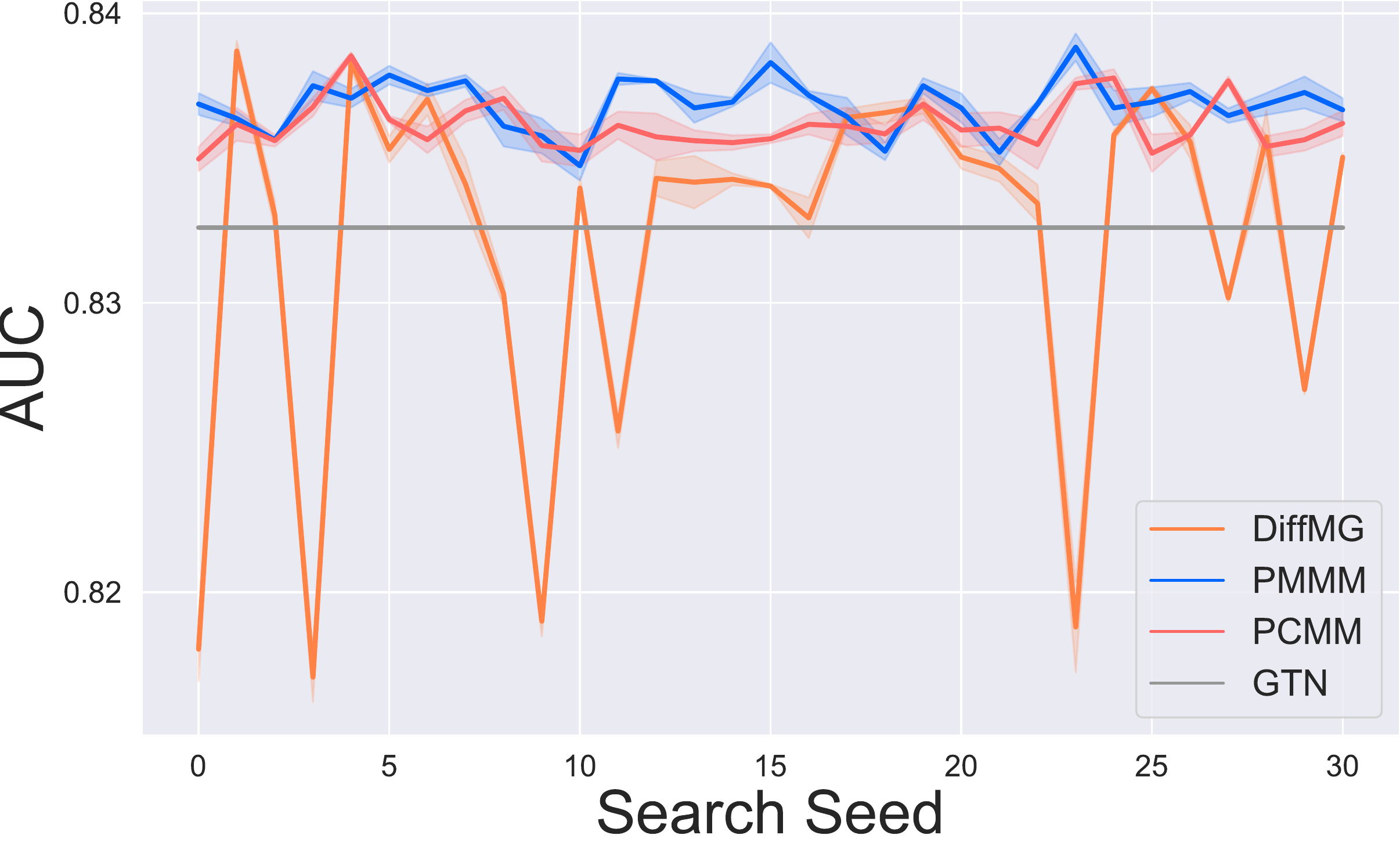}}
\caption{Performance on different search seeds (best viewed in color). The first three figures are results for the node classification task, followed by three figures for the recommendation task.
}
\vspace{-1ex}
\label{fig:seed}
\end{figure*}

\subsection{Consistency}
An essential issue of differential search is the gap between the super-net in the search stage and the target-net in the evaluation stage, which means the target-net obtained from a high-performance super-net is not guaranteed to be good~\cite{yang2020ista,yang2021towards}.  
To show that our search algorithm on \model~can reduce the inconsistency between super-net and target-net, we adopt the Kendall metric~\cite{kendall1938new} to evaluate the rank correlation of result pairs. Specifically, we run DARTS~\cite{DBLP:conf/iclr/LiuSY19}, DiffMG~\cite{DBLP:conf/kdd/DingYZZ21}, \mone~and \model~on DBLP, IMDB, and ACM for six times with different search seeds, and retrain the searched target-net. We employ the six pairs of super-net and target-net accuracy of each method to calculate the Kendall metric on each dataset.
When the ranking order changes from reversed to identical, the Kendall metric varies from $-1$ to $1$, and a higher Kendall metric indicates better consistency. As shown in Table~\ref{tab:consistent}, DARTS shows the lowest Kendall metric scores because it utilizes all candidate paths in each iteration of the search stage while only deriving one path in the evaluation stage, resulting in a significant gap. Because of simpling partial paths in the search stage, DiffMG and \mone~have higher Kendall metric scores. However, they still contain deriving process. 
\model~shows the highest Kendall metric scores on all three datasets, indicating the effectiveness of the progressive algorithm for improving consistency. Therefore, a higher super-net performance is more likely to indicate a better target-net architecture.

\begin{table}[!t]
	\centering
	\caption{Comparision of Kendall metric with related methods.}
	\label{tab:consistent}
	\begin{threeparttable}[b]
    \footnotesize
    \resizebox{0.48\textwidth}{!}{
	\begin{tabular}{p{1cm}<{\centering}
	p{1cm}<{\centering} p{1.5cm}<{\centering} p{1cm}<{\centering}
	}
		\toprule
		Methods &   IMDB              & DBLP                     & ACM   \\ 
		\midrule
		DARTS    & -0.20  &  0.00 & 0.07  \\
		DiffMG   & 0.20   &  0.23 & 0.07  \\
		\mone   & 0.07  &  0.55 & 0.20  \\
		\model   & \textbf{0.73}  &  \textbf{0.60} & \textbf{0.47}  \\
		
        \bottomrule
	\end{tabular}}
	\end{threeparttable}
\end{table}

\subsection{Stability}

\label{sec:robust}

To evaluate the stability of our method, we compare \mone~and \model~with differentiable meta-graph search by using different random search seeds. 
We run the three algorithms on random search seeds from $0$ to $30$, and plot the Macro-F1 and AUC scores averaged from $3$ dependent retraining of the searched architecture under different random training seeds. The results are illustrated in Figure~\ref{fig:seed}.
The gray dotted line shows the results of hand-designed heterogeneous GNNs, HAN on node classification, and GTN on recommendation as the baselines. Although DiffMG shows excellent performance in a few search seeds, the performance dramatically declines in most other seeds. In most cases, its performance is even worse than HAN and GTN.  
In contrast, both \mone~and \model~can overcome the instability issue in DiffMG. Besides, \mone~and \model~significantly outperform DiffMG in most search seeds and consistently surpass the manually designed networks. Note that although \model~is designed for improving consistency, it has similar stability to \mone~because inconsistency is one of the essential reasons for instability. Consequently, PCMM is a more advanced algorithm than PMMM.  





\begin{figure*}[!t]
\centering
\subfloat[DBLP]{\includegraphics[width=0.45\linewidth]{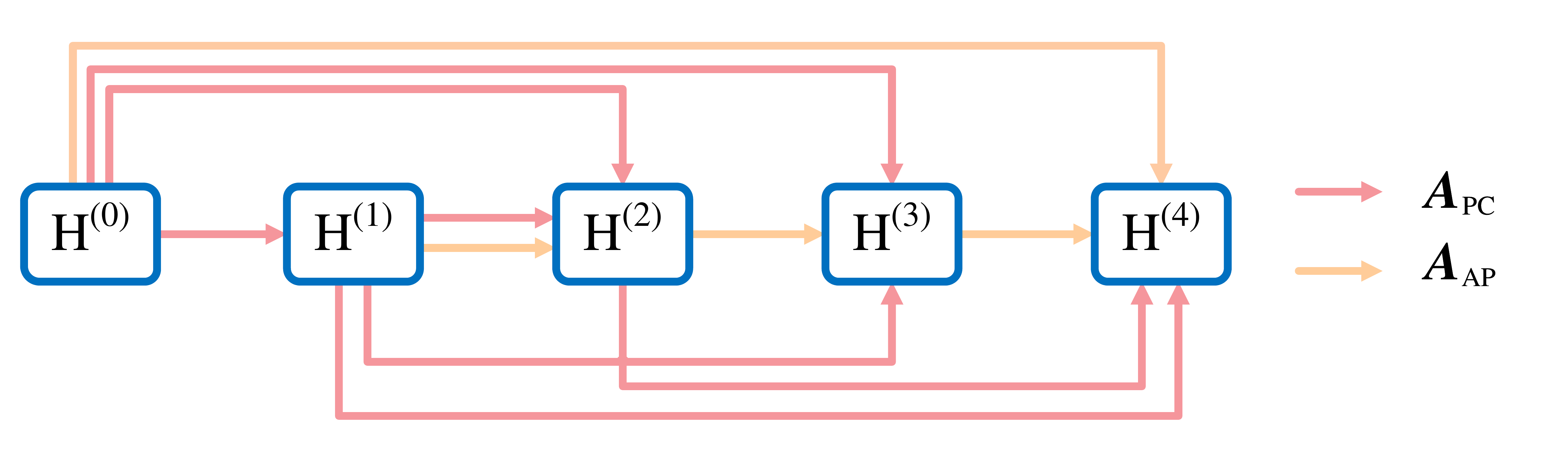}}
\hspace{1.5em}
\subfloat[DBLP]{\includegraphics[width=0.45\linewidth]{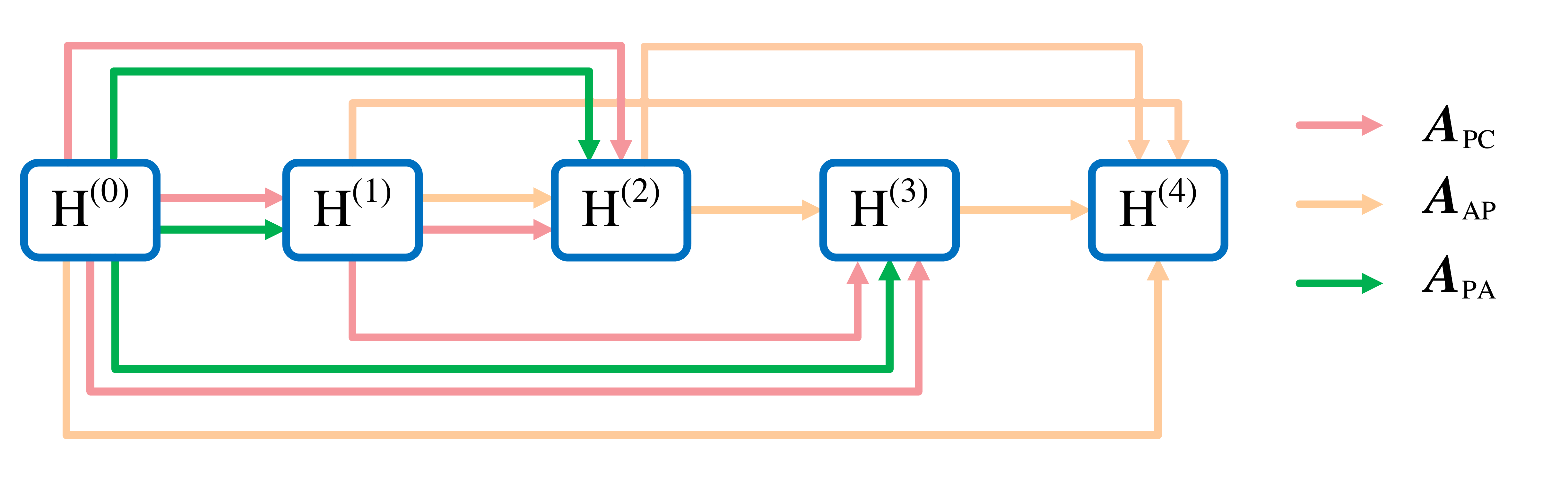}}
\\
\subfloat[Yelp]{\includegraphics[width=0.90\linewidth]{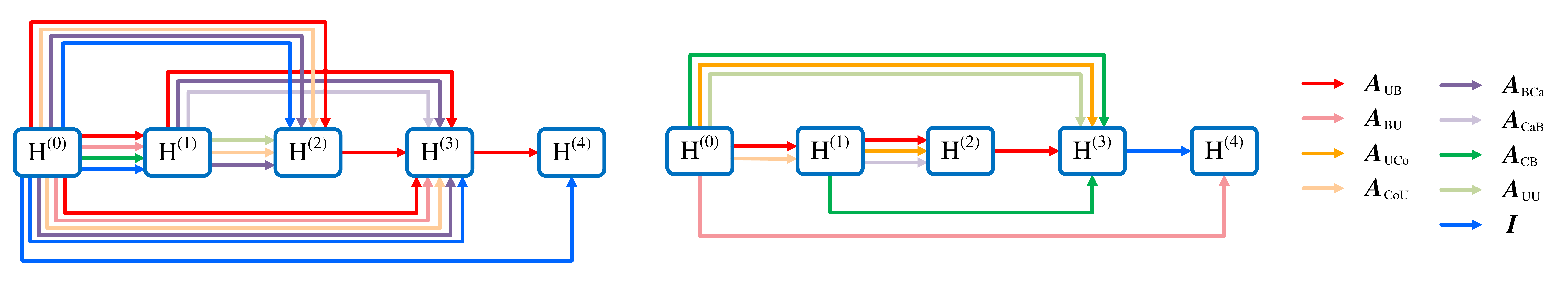}}

\caption{Meta-multigraphs searched by \mone~(a) and \model~(b), (c). Each edge type is associated with a simple mean aggregator, so the final meta-structures-based HGNNs are efficient.}

\label{fig:visual}
\end{figure*}
\begin{figure*}[!t]
\centering
\subfloat[GCN]{\includegraphics[width=0.28\linewidth]{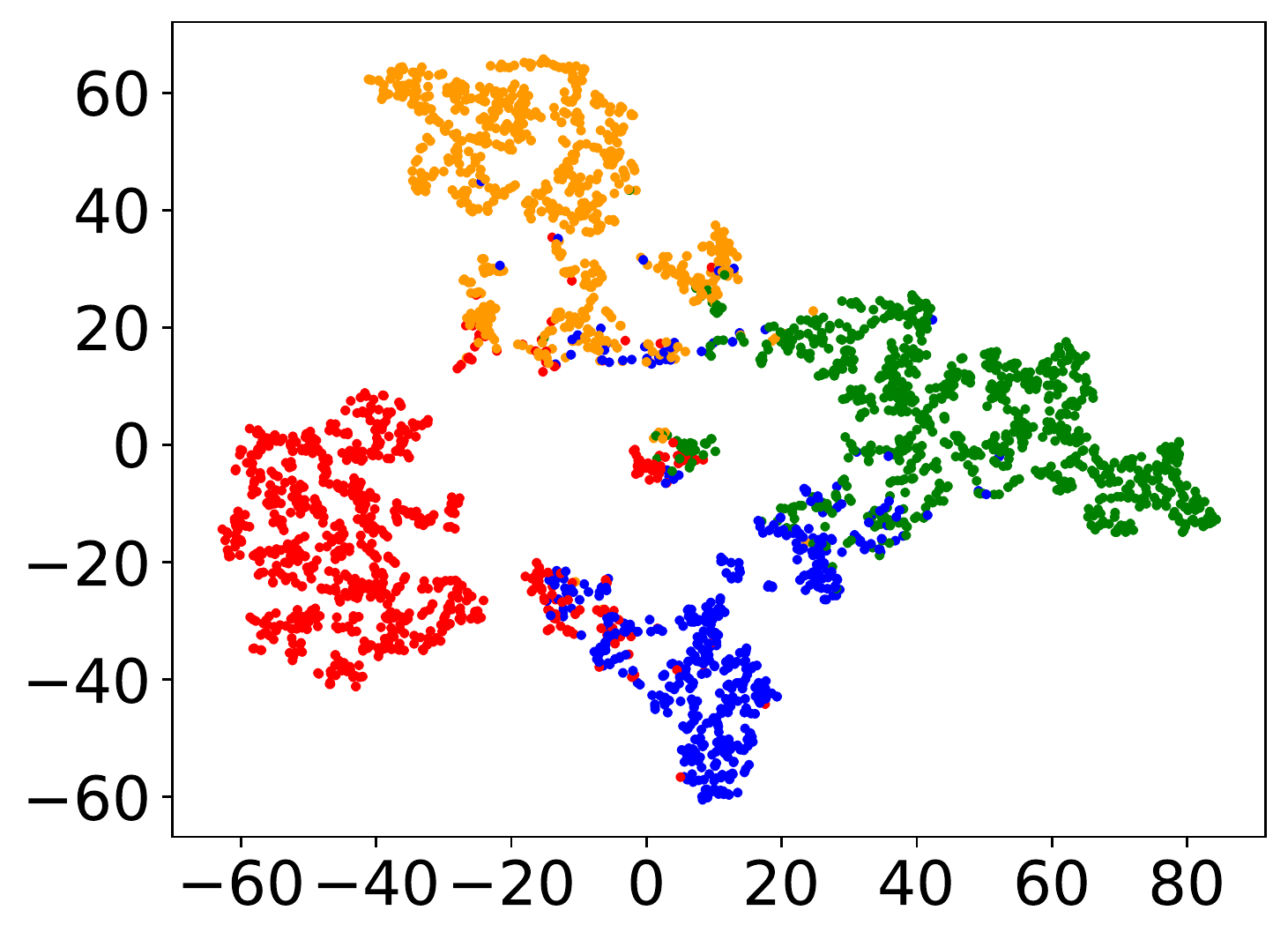}}
\hspace{2em}
\subfloat[HAN]{\includegraphics[width=0.28\linewidth]{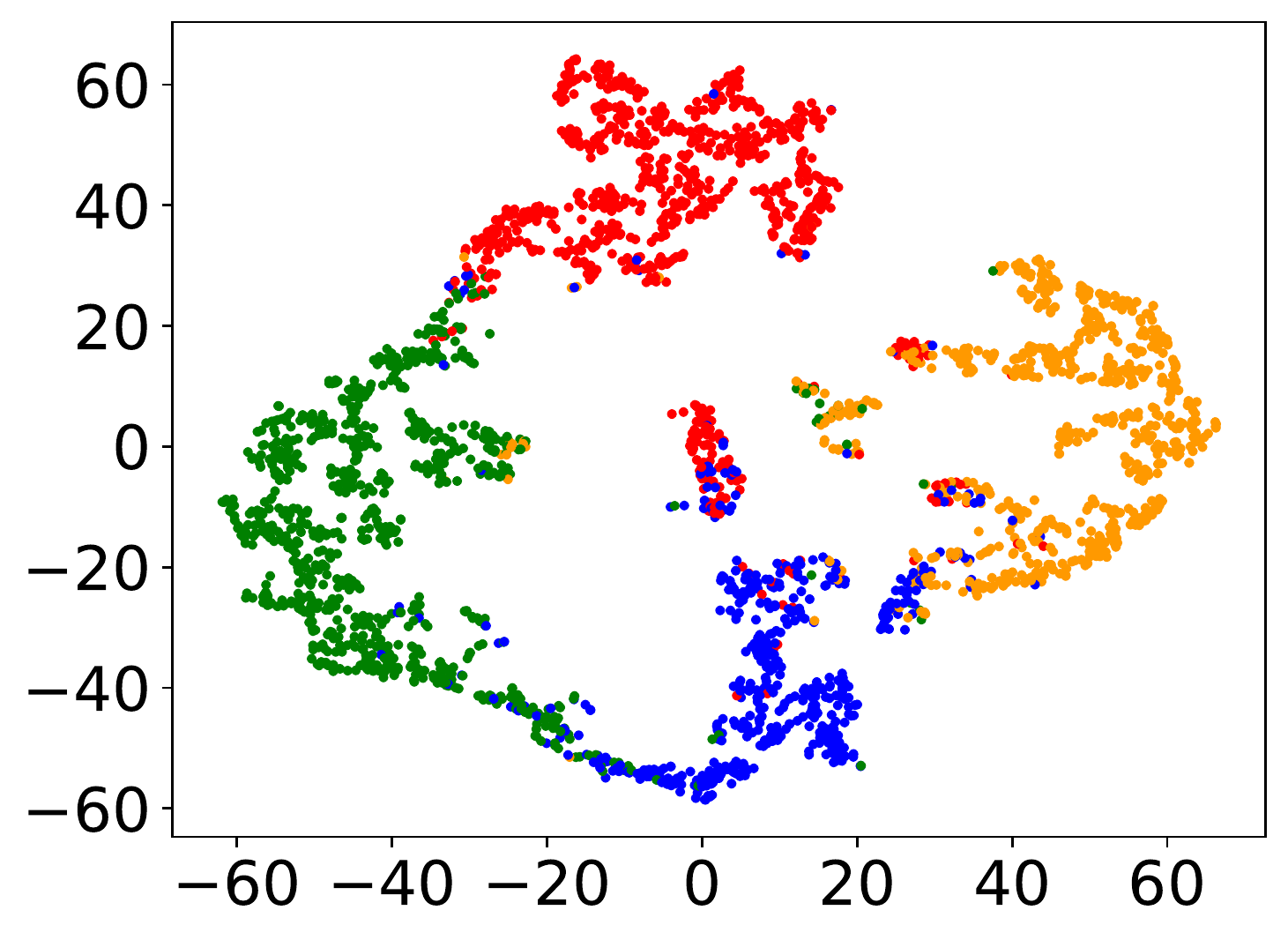}}
\hspace{2em}
\subfloat[Simple-HGN]{\includegraphics[width=0.28\linewidth]{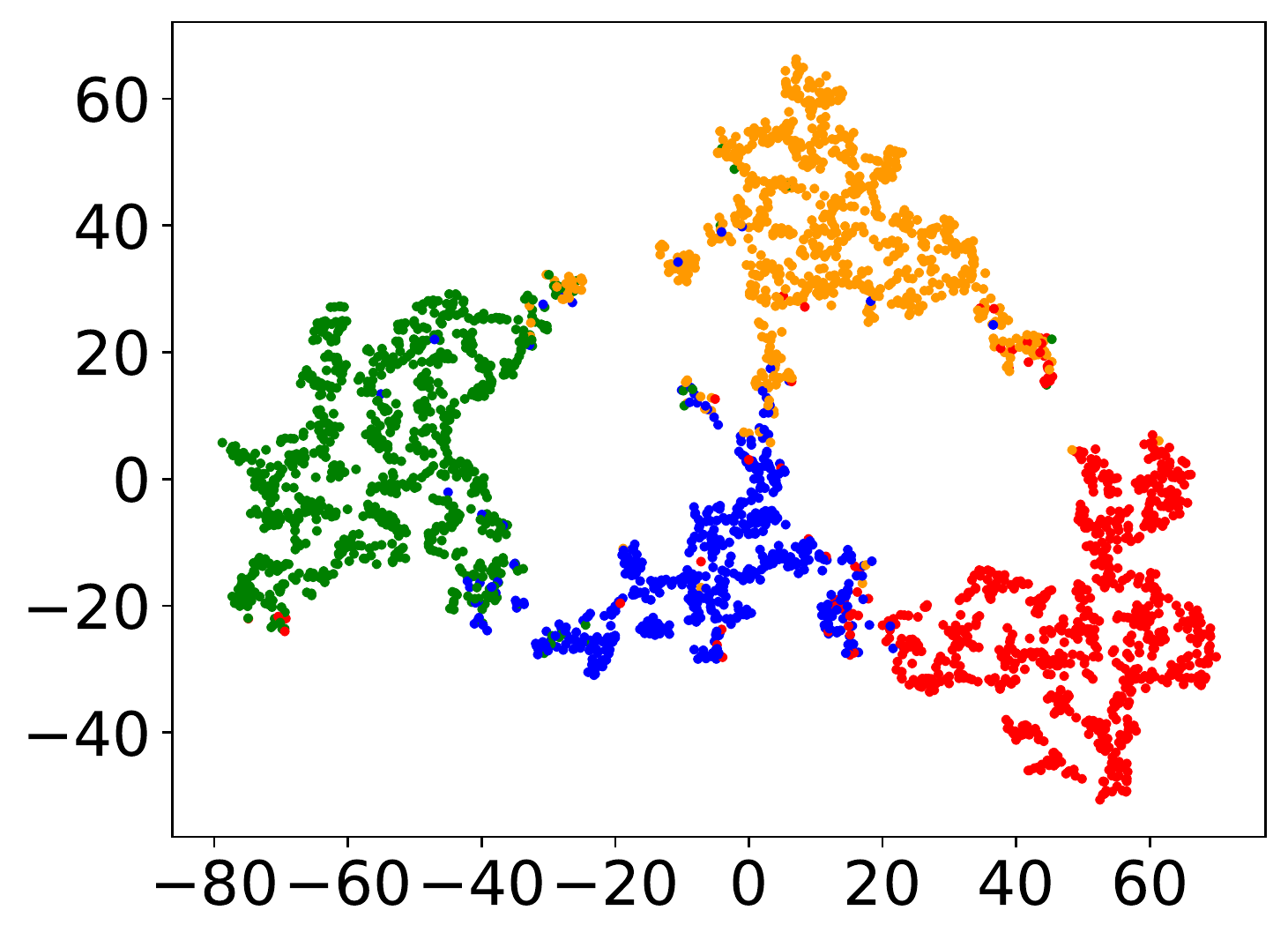}}

\subfloat[DiffMG]{\includegraphics[width=0.28\linewidth]{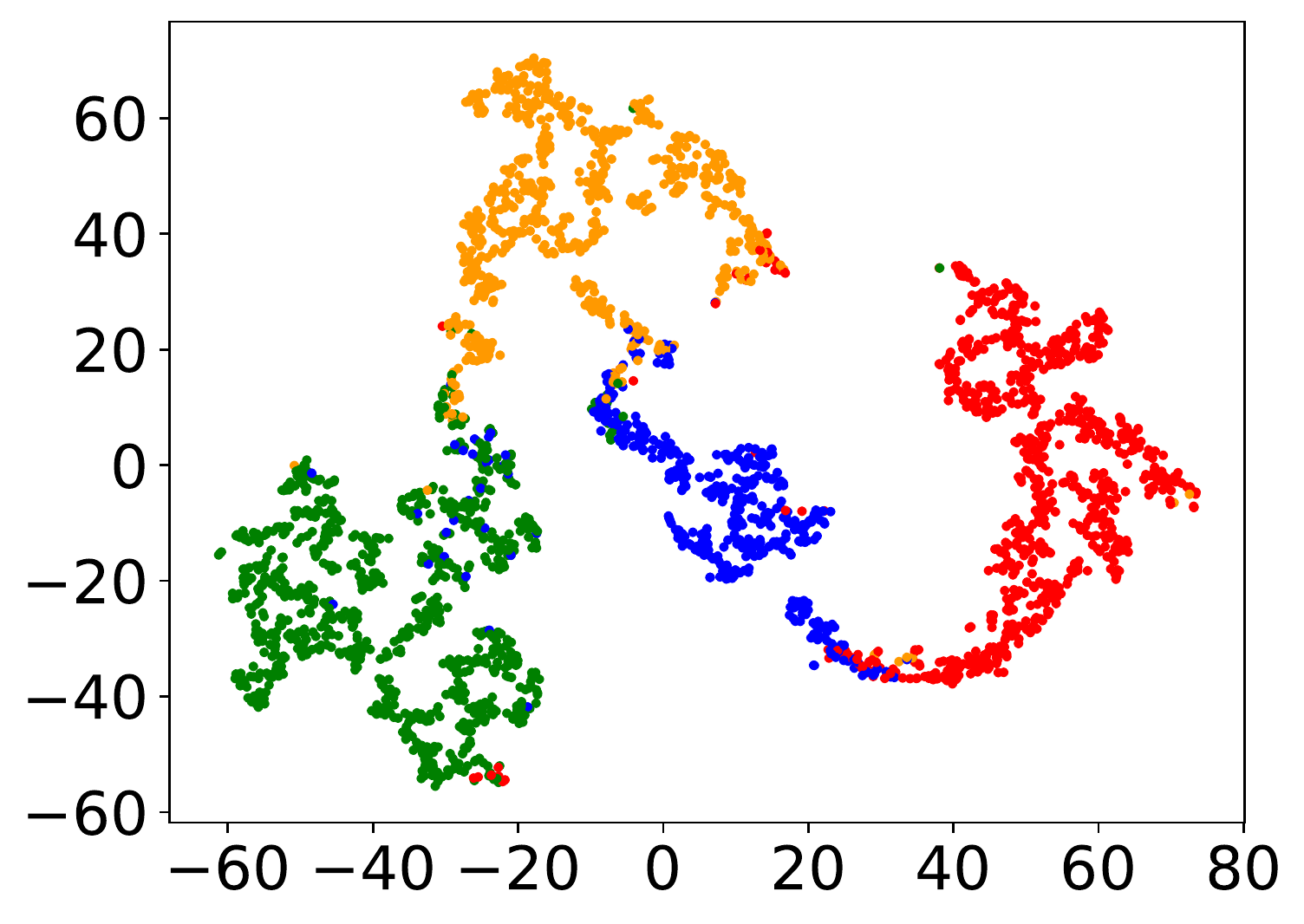}}
\hspace{2em}
\subfloat[PMMM]{\includegraphics[width=0.28\linewidth]{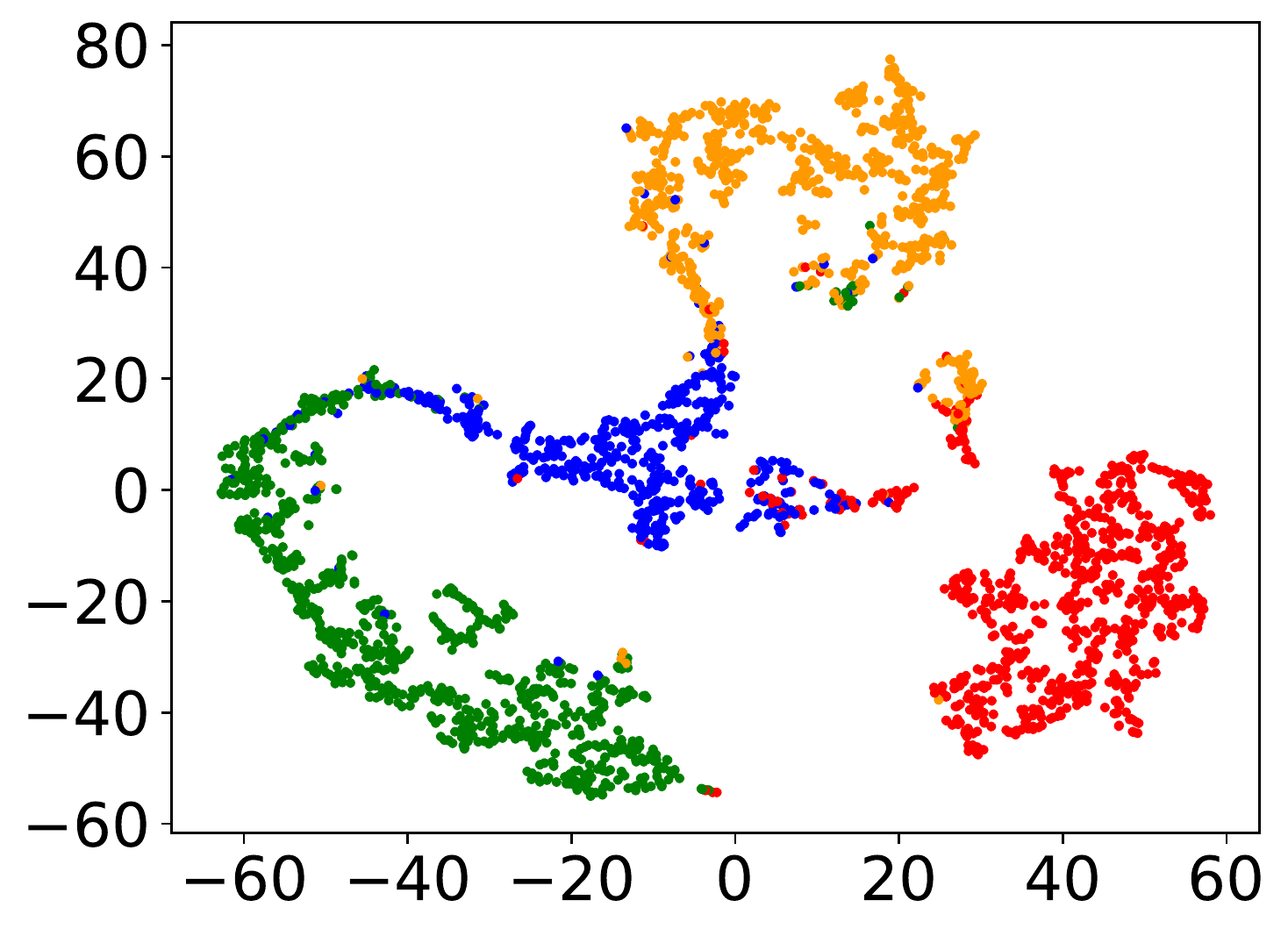}}
\hspace{2em}
\subfloat[\model]{\includegraphics[width=0.28\linewidth]{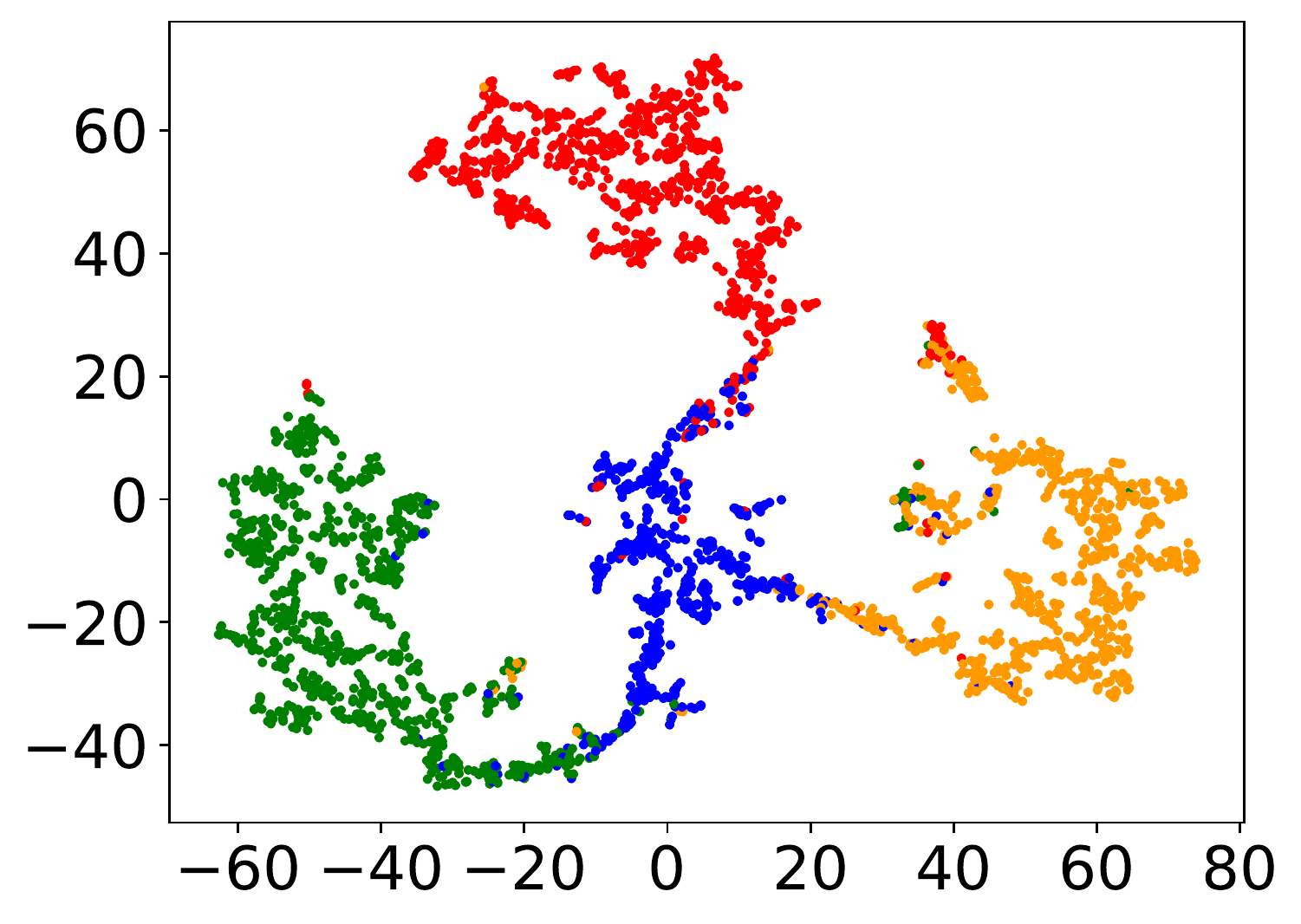}}
\caption{Visualization embedding on DBLP. Each point indicates one author, and its color indicates the research area. 
}
\vspace{-1ex}
\label{fig:visualization}
\end{figure*}
\subsection{Visualization of Searched Meta-multigraphs}

Fig.~\ref{fig:visual} visualizes the meta-multigraphs searched by \mone~and \model~on DBLP for node classification and \model~on Amazon for recommendation task. The searched meta-structures by \mone~and \model~are more complex than those of DiffMG. However, more message-passing steps have a slightly negative impact on efficiency due to the use of the simple mean aggregator and parallelization in the training of neural architecture, which has been verified by Fig.~\ref{fig:efficiency}. In addition, both meta-multigraphs searched by \model~have complex message-passing steps in the shallow parts while concise message-passing steps in the deep parts, indicating more exploration in the shallow parts while more exploitation in the deep parts.
In our experiments, we find that both $\bm{\mathcal{A}}_{PC}$ and $\bm{\mathcal{A}}_{PA}$ are critical for message-passing between $\bm{H}^{(0)}$ and $\bm{H}^{(2)}$ in Fig.~\ref{fig:visual} (b). If we search for a meta-graph like DiffMG, $\bm{\mathcal{A}}_{PA}$ has to be discarded as multiple paths are not allowed, which will seriously 
affect performance and cause degradation. So a complex meta-structure, \eg, meta-multigraph, is more expressive than a simple meta-structure, \eg, meta-path or meta-graph.

\subsection{Visualization of Node Embedding}
To enhance the comprehensibility of our comparison, we perform a visualization task that aims to lay out a heterogeneous graph in a low-dimensional space. Specifically, we learn the node embedding based on the proposed model and subsequently project these embeddings onto a 2-dimensional space. To facilitate visualization of our results, we utilize t-SNE~\cite{maaten2008visualizing} to visualize the author embedding in DBLP, with each node being assigned a color based on its respective research domain.

From Figure~\ref{fig:visualization}, one can see that GCN and HAN don't perform well because the authors belonging to different research areas are mixed with each other. 
DiffMG performs much better, demonstrating that the searched meta-graph can significantly contribute to heterogeneous graph analysis. However, since DiffMG only considers meta-graphs, the boundary is still blurry. For example, the boundary between green nodes and yellow nodes is blurry. In contrast, the boundaries between green, yellow, and red nodes are more distinct
in PMMM and \model, especially in \model. With the guide of complex-to-concise meta-multigraph, the embedding learned by \model~has high intra-class similarity, and it separates the authors in the different research areas with distinct boundaries.

\subsection{Ablation Studies on Critical Modules}


\subsubsection{Comparison among Different Meta-structures}
\begin{table}[!t]
	\centering
	\caption{Macro-F1 ($\%$) of ablation study on different meta-structures. 
 }
	\label{tab:meta-structure}
	\begin{threeparttable}[b]
    \footnotesize
    \resizebox{0.46\textwidth}{!}{
	\begin{tabular}{lccc 
	}
		\toprule
		&    IMDB               &  DBLP                      &  ACM   \\ 
		\midrule
		Meta-path  & $ 61.81_{\pm0.59      } $ & $ 94.62_{\pm0.18 } $ & $ 92.13_{\pm0.34} $ \\
		Meta-graph  & $  63.31_{\pm0.54    } $ & $ 94.70_{\pm0.16 } $ & $ 92.86_{\pm0.39} $ \\
		MM  & $ 64.32_{\pm0.35 } $ & $ 94.72_{\pm0.10 
        } $ & $ 93.34_{\pm0.41} $  \\
		\midrule
		\meta~MM  & $  \bm{64.92_{\pm0.23   }} $ & $ \bm{94.86_{\pm0.13       }} $ & $ \bm{94.02_{\pm0.21}} $ \\
        \bottomrule
\end{tabular}}
\begin{tablenotes}
\footnotesize
\item MM: Meta-multigraph.
\end{tablenotes}
\end{threeparttable}
\end{table}
To explore the relationship between the performance of HGNNs and the types of searched meta-structures, we search for four different meta-structures based on Section~\ref{sec:meta}. The results on IMDB, DBLP, and ACM are illustrated in Table~\ref{tab:meta-structure}. Meta-path shows the worst performance, indicating the simple one path is insufficient for modeling message propagation of HINs. Meta-graph exhibits better performance than meta-path, and meta-multigraph outperforms meta-graph on all three datasets. So we can conclude that the performance improves along with the complexity of the searched meta-structures. \meta~meta-multigraph is a variant of meta-multigraph. It has a similar complexity to meta-multigraph. However, \meta~meta-multigraph propagates information from complex to concise, making it more expressive. So \meta~meta-multigraph outperforms the other three meta-structures. 

\subsubsection{Effectiveness of Search Algorithm and Meta-structure}
\begin{table}[!t]
	\centering
	\caption{Macro-F1 ($\%$) of ablation study results on the critical modules of PMMM and~\model.}
	\label{tab:part}
	\begin{threeparttable}[b]
    \footnotesize
    \resizebox{0.48\textwidth}{!}{
	\begin{tabular}{lccc 
	}
		\toprule
		&    IMDB              &  DBLP                    & ACM   \\ 
		\midrule
		DiffMG  & $ 49.35_{\pm1.89} $ & $  93.68_{\pm0.37}$ & $  88.03_{\pm3.25}$ \\
		DiffMG + Partial message & $ 63.02_{\pm0.46 }$ & $  94.55_{\pm0.32} $ & $  92.66_{\pm0.31}$ \\
		DiffMG + MM  & $ 63.65_{\pm0.53 }$ & $ 94.38_{\pm0.25} $ & $ 92.73_{\pm0.19}$ \\
		\midrule
		PMMM & $ \bm{64.49_{\pm0.39 }} $ & $ \bm{94.79_{\pm0.14 }} $ & $  
         \bm{93.79_{\pm0.16}}$ \\
		\midrule
		\midrule
		DiffMG + Progressive & $ 63.12_{\pm0.60 }$ & $  94.70_{\pm0.16} $ & $  92.86_{\pm0.39}$ \\
		DiffMG + \meta~MM  & $ 63.71_{\pm0.74 }$ & $ 94.74_{\pm0.10} $ & $ 92.98_{\pm0.14}$ \\
		\midrule
		\model  & $ \bm{64.92_{\pm0.23 }} $ & $ \bm{94.86_{\pm0.13 }} $ & $  \bm{94.02_{\pm0.21}}$ \\
        \bottomrule
	\end{tabular}}
 \begin{tablenotes}
\footnotesize
\item MM: Meta-multigraph.
\end{tablenotes}
	\end{threeparttable}
\vspace{-1ex}
\end{table}

In this work, we propose two search algorithms, PMMM and \model. PMMM employs a partial message strategy to search for meta-multigraphs, while \model~uses a progressive algorithm to search for \meta~meta-multigraphs. Here, we explore how each of them improves the performance through ablation studies on IMDB, DBLP, and ACM. Specifically, we separately combine the partial message strategy and meta-multigraphs with DiffMG, and compare them with PMMM.
We also separately combine the progressive algorithm and \meta~meta-multigraph with DiffMG and compare them with \model. 
The results are shown in Table~\ref{tab:part}. Even if only one of our designs is employed, the improvement compared to DiffMG is considerable, which verifies that our designs are quite effective.   
Furthermore, the progressive algorithm performs marginally better than the partial message strategy, which verifies the effectiveness of closing the inconsistency. \meta~meta-multigraphs also have better performance than meta-multigraphs, demonstrating the value of eliminating 
redundant paths.


\subsection{Hyper-parameter Study}
\begin{table}[!t]
\centering
\caption{Ablation study on sampling proportion $p$ in PMMM.}
\label{tab:p}
\begin{threeparttable}[b]
\footnotesize
\resizebox{0.48\textwidth}{!}{
\begin{tabular}{lcccc}
\toprule
&        $p$ = 1     &       $p$ = 2          &   $p$ = 3 &$p$ = 4\\ 
\midrule
IMDB& $64.21_{\pm0.29}$      & $\bm{64.49_{\pm0.39}}$ & $64.37_{\pm0.27}$ & $64.39_{\pm0.25}$\\
DBLP& $\bm{94.79_{\pm0.14}}$     & $94.76_{\pm0.17}$ & $94.68_{\pm0.23}$ & $94.66_{\pm0.14}$ \\
ACM& $\bm{93.79_{\pm0.16}}$  &  $93.48_{\pm0.21}$      & $93.66_{\pm0.15}$ & $93.56_{\pm0.11}$\\
Amazon& $75.22_{\pm0.08}$  &  $\bm{75.48_{\pm0.04}}$      & $75.39_{\pm0.12}$ &$75.44_{\pm0.05}$\\
Yelp& $68.98_{\pm0.16}$  &  $69.27_{\pm0.13}$      & $\bm{69.29_{\pm0.17}}$ & $69.17_{\pm0.21}$\\
Douban& $83.39_{\pm0.06}$  &  $\bm{83.88_{\pm0.02}}$      & $83.68_{\pm0.07}$ & $83.83_{\pm0.04}$\\
\bottomrule
\end{tabular}}
\end{threeparttable}
\end{table}

\subsubsection{Ablation Study on Sampling Proportion}
In PMMM, we sampled $1/p$ paths in each multi-edge to overcome the coupling optimization. We perform ablation studies on hyper-parameter $p$ in Table~\ref{tab:p}. For most datasets, PMMM performs better when $p>1$ than $p=1$, which means the coupling optimization will hurt the performance, and the partial message strategy is useful. The coupling optimization does not affect DBLP and ACM because the number of candidate paths on the two datasets is small. PMMM performs similarly when $p>1$, so we set $p=2$ for convenience.

\begin{table}[!t]
\centering
\caption{Ablation study on attenuation ratio $\beta$ in \model.}
\label{tab:beta}
\begin{threeparttable}[b]
\footnotesize
\resizebox{0.48\textwidth}{!}{
\begin{tabular}{lcccc}
\toprule
&        $\beta$ = 1     &       $\beta$ = 0.9          &   $\beta$ = 0.7 &$\beta$ = 0.5\\ 
\midrule
IMDB& $63.66_{\pm0.19}$      & $\bm{64.92_{\pm0.39}}$ & $64.30_{\pm0.21}$ & $64.11_{\pm0.27}$\\
DBLP& $94.37_{\pm0.21}$     & $94.86_{\pm0.14}$ & $\bm{94.88_{\pm0.21}}$ & $94.02_{\pm0.16}$ \\
ACM& $93.42_{\pm0.12}$  &  $\bm{94.02_{\pm0.16}}$      & $93.60_{\pm0.11}$ & $93.51_{\pm0.13}$\\
Amazon& $75.03_{\pm0.09}$  &  $\bm{75.55_{\pm0.04}}$      & $75.29_{\pm0.17}$ &$75.40_{\pm0.07}$\\
Yelp& $68.88_{\pm0.13}$  &  $\bm{69.54_{\pm0.17}}$      & $69.20_{\pm0.13}$ & $69.11_{\pm0.22}$\\
Douban& $83.33_{\pm0.07}$  &  $\bm{83.99_{\pm0.02}}$      & $83.61_{\pm0.06}$ & $83.80_{\pm0.05}$\\
\bottomrule
\end{tabular}}
\end{threeparttable}
\end{table}
\subsubsection{Ablation Study on Attenuation Ratio}
To generate a \meta~meta-multigraph, we need to adjust the threshold $\tau^{(i,j)}$ for different depths based on attenuation ratio $\beta$. We perform ablation studies on hyper-parameter $\beta$ in Table~\ref{tab:beta}. When $\beta=1$, only the strongest path is retained in each edge, which is the deriving strategy of most differentiable NAS. The smaller $\beta$ is, the more paths are retained. We speculate that $\beta$ should be close to 1 to ensure that all effective paths are retained and weak paths are dropped. As shown in Table~\ref{tab:beta}, \model~shows the best performance at $\beta=0.9$, verifying our speculation. 

\begin{table}[!t]
	\centering
	\caption{Macro-F1 ($\%$) of ablation study on grow strategy of $\lambda_e$.}
	\label{tab:lambda}
	\begin{threeparttable}[b]
    \footnotesize
    \resizebox{0.48\textwidth}{!}{
	\begin{tabular}{lccc 
	}
		\toprule
		\textit{Warm-up} &   IMDB              & DBLP                     & ACM   \\ 
		\midrule
		$0$ epoch (ours) & $64.92_{\pm0.23}$      & $\bm{94.79_{\pm0.13}}$ & $\bm{94.02_{\pm0.21}}$ \\
		$5$ epochs & $62.51_{\pm0.37}$     & $94.83_{\pm0.18}$ & $92.66_{\pm0.24}$ \\
		$10$ epochs & $62.56_{\pm0.28}$  &  $93.21_{\pm0.25}$      & $93.21_{\pm0.25}$ \\
        \midrule
        \textit{Quadratically} & $\bm{64.94_{\pm0.18}}$  &  $94.61_{\pm0.21}$      & $93.91_{\pm0.23}$ \\
        \textit{Step} & $63.96_{\pm0.28}$  &  $94.70_{\pm0.19}$      & $93.98_{\pm0.16}$ \\
		\bottomrule
	\end{tabular}}
	\end{threeparttable}
\end{table}
\subsubsection{Ablation Study on Grow Strategy}
We use a straightforward grow strategy for $\lambda_e$, which grows linearly as the epoch increases. Intuitively, we should warm up the architecture parameters for several epochs before increasing $\lambda_e$ to perform the path selection. 
So we design two extra strategies for comparison. Specifically, $\lambda_e$ is fixed to $0$ at the first $n~(n=5,10)$ epochs and then increases, which means the super-net does not perform path selection in the first $n$ epochs.  
However, as shown in Table~\ref{tab:lambda}, warm-up strategies are not as beneficial as we expected. We speculate that it is because our progressive algorithm is performed end-to-end. 
Consequently, even if the candidate paths inactivated early are not unreasonable as architecture parameters are not fully trained, they could be reactivated later when weaker paths appear. On the contrary, warm-up strategies will introduce extra coupling optimization issue~\cite{guo2020single}. Except \textit{linear} and \textit{warm-up}, we also design two extra strategies for comparison: \textit{quadratically} increasing and \textit{step} increasing. They both show similar
performance. Hence, our method is insensitive to the type of Grow policy on $\lambda_e$.

\section{Conclusion}
\label{sec.Conclusion}
In this work, we explore the relationship between the performance of HGNNs and types of meta-structures, and 
find that existing meta-structures are not ideal for encoding various semantic information on diverse heterogeneous information networks. 
Therefore, we introduce a new concept called meta-multigraph, which is a more expressive and flexible generalization of a meta-graph. To search for a meaningful meta-multigraph for specific datasets or tasks, 
we propose a partial message search algorithm for solving the instability issue of meta-multigraph search. Because the flexibility of meta-multigraphs may result in the propagation of redundant messages, we introduce a complex-to-concise meta-multigraph. Then, we propose a more advanced progressive algorithm to search for complex-to-concise meta-multigraphs, which further overcomes the inconsistent issues between the meta-structures in search and evaluation.   
Extensive experiments on two representative HIN tasks 
demonstrate that both of our methods can consistently find effective meta-structures and achieve state-of-the-art performance on various datasets.



\printcredits

\section*{Acknowledgments}
This work is supported by the National Natural Science Foundation of China (No. U22B2017).

\bibliographystyle{unsrt}
\bibliography{cas-refs}


\begin{highlights}
\item Research highlights item 1
\item Research highlights item 2
\item Research highlights item 3
\end{highlights}

\end{document}